\documentclass[lettersize,journal]{IEEEtran}
\usepackage{amsmath,amsfonts}
\usepackage{amssymb}
\usepackage{algorithmic}
\usepackage{algorithm}
\usepackage{array}
\usepackage{subfigure}
\usepackage{textcomp}
\usepackage{stfloats}
\usepackage{url}
\usepackage{verbatim}
\usepackage{graphicx}
\usepackage{booktabs}
\usepackage{multirow}
\usepackage{color,xcolor}
\usepackage{cite}

\usepackage{url}

\usepackage{pifont}
\begin{document}

\title{Spectral Discrepancy and Cross-modal Semantic Consistency Learning for Object Detection in Hyperspectral Images}

\author{Xiao He, 
	Chang Tang, ~\IEEEmembership{Senior Member,~IEEE,}
 Xinwang Liu, ~\IEEEmembership{Senior Member,~IEEE,}
 Wei Zhang,
         Zhimin Gao,
         Chuankun Li,
         Shaohua Qiu,
        Jiangfeng Xu
    
	\IEEEcompsocitemizethanks{
  \IEEEcompsocthanksitem This work was supported in part by the Natural Natural Science Foundation of Shandong Province under grant ZR2021LZH001, and in part by the National Natural Science Foundation of China under grant 62476258, and in part by the Natural Science Foundation of Hubei Province under grant 2025AFA113. (\textit{Corresponding author: Chang Tang}.)
 \IEEEcompsocthanksitem X. He is with the School of Computer, Wuhan University, Wuhan, China. E-mail: xiaohewhu@163.com.
\IEEEcompsocthanksitem C. Tang is with the School of Software Engineering, Huazhong University of Science and Technology, Wuhan 430074, China. E-mail: tangchang@hust.edu.cn.
\IEEEcompsocthanksitem X. Liu is with the school of computer, National University of Defense Technology, Changsha 410073, China. E-mail: xinwangliu@nudt.edu.cn.
\IEEEcompsocthanksitem W. Zhang is with Shandong Provincial Key Laboratory of Computer Networks, Shandong Computer Science Center (National Supercomputing Center in Jinan), Qilu University of Technology (Shandong Academy of Sciences), Jinan 250000, China. E-mail:  wzhang@qlu.edu.cn.
 \IEEEcompsocthanksitem Z. Gao is with the School of Computer and Artificial Intelligence, Zhengzhou University, Zhengzhou 450001, China. E-mail: zhimingao113@gmail.com.
 \IEEEcompsocthanksitem C. Li is currently with the School of Information and Communication Engineering, North University of China, Taiyuan, China, E-mail: chuankun@nuc.edu.cn.
\IEEEcompsocthanksitem S. Qiu is with the National Key Laboratory of Electromagnetic Energy, Naval University of Engineering, Wuhan 430030, China. E-mail: qiush125@163.com.
\IEEEcompsocthanksitem J. Xu is with the Hexagon AB. E-mail: jiangfeng.xu@hexagon.com.
}
 
 }

\markboth{IEEE Transactions on Multimedia}%
{Shell \MakeLowercase{\textit{et al.}}: A Sample Article Using IEEEtran.cls for IEEE Journals}


\maketitle

\begin{abstract}
Hyperspectral images with high spectral resolution provide new insights into recognizing subtle differences in similar substances. However, object detection in hyperspectral images faces significant challenges in intra- and inter-class similarity due to the spatial differences in hyperspectral inter-bands and unavoidable interferences, e.g., sensor noises and illumination. To alleviate the hyperspectral inter-bands inconsistencies and redundancy, we propose a novel network termed \textbf{S}pectral \textbf{D}iscrepancy and \textbf{C}ross-\textbf{M}odal semantic consistency learning (SDCM), which facilitates the extraction of consistent information across a wide range of hyperspectral bands while utilizing the spectral dimension to pinpoint regions of interest. Specifically, we leverage a semantic consistency learning (SCL) module that utilizes inter-band contextual cues to diminish the heterogeneity of information among bands, yielding highly coherent spectral dimension representations. On the other hand, we incorporate a spectral gated generator (SGG) into the framework that filters out the redundant data inherent in hyperspectral information based on the importance of the bands. Then, we design the spectral discrepancy aware (SDA) module to enrich the semantic representation of high-level information by extracting pixel-level spectral features. Extensive experiments on two hyperspectral datasets demonstrate that our proposed method achieves state-of-the-art performance when compared with other ones.
\end{abstract}

\begin{IEEEkeywords}
 	hyperspectral, multi-modal, feature fusion, object
   detection, consistency learning.
\end{IEEEkeywords}
	\section{Introduction}

 \IEEEPARstart{W}{ith} the advancement of spectral imaging technology, optical remote sensing has embraced diverse imaging modalities including RGB-T~\cite{he2023multispectral}, multispectral, and hyperspectral imagery~\cite{sun2019hyperspectral}, thus extending its coverage across a broader electromagnetic spectrum range~\cite{ghamisi2017advances}. Among these modalities, hyperspectral imagery, unlike conventional visible images, comprises sequential bands, with each capturing distinct spectral information~\cite{9709645}. Benefitting from this intricate spectral profile and supported by the advancements in imaging and image classification, hyperspectral data has found applications in various domains, such as salient object detection~\cite{liang2018material}, medical image processing~\cite{lu2014medical}, mineral exploration~\cite{bedini2017use}, and semantic segmentation~\cite{lu2020recent}.

In the interpretation and analysis of hyperspectral images, object detection plays a crucial role in extracting valuable information from specific regions of interest~\cite{yan2021object}. Generally, object detection methods for visible images mainly leverage texture and shape features from the spatial dimension to localize the object. In contrast, hyperspectral images not only contain spatial information but also provide rich spectral information in the multiple wavelength bands, which can be utilized to enhance the detection performance~\cite{liang2018material,10313066}. On the other hand, object detection in hyperspectral images presents significant challenges, with the primary hurdle lying in how to effectively exploit the intricate spatial and spectral information~\cite{yu2020simplified, kumar2023methanemapper}. Besides, due to the high cost of hyperspectral data storage and complex preprocessing, there are a few existing hyperspectral images based object detection. Yan et al.\cite{yan2021object} designed a 3D convolution framework to capture spatial and spectral relationships and developed the first dataset for hyperspectral images based object detection. Building upon DETR\cite{dai2021dynamic}, Satish et al.~\cite{kumar2023methanemapper} utilized spectral dimension properties to guide detection framework.

\begin{figure}
	\centering
	\includegraphics[width = 1\linewidth]{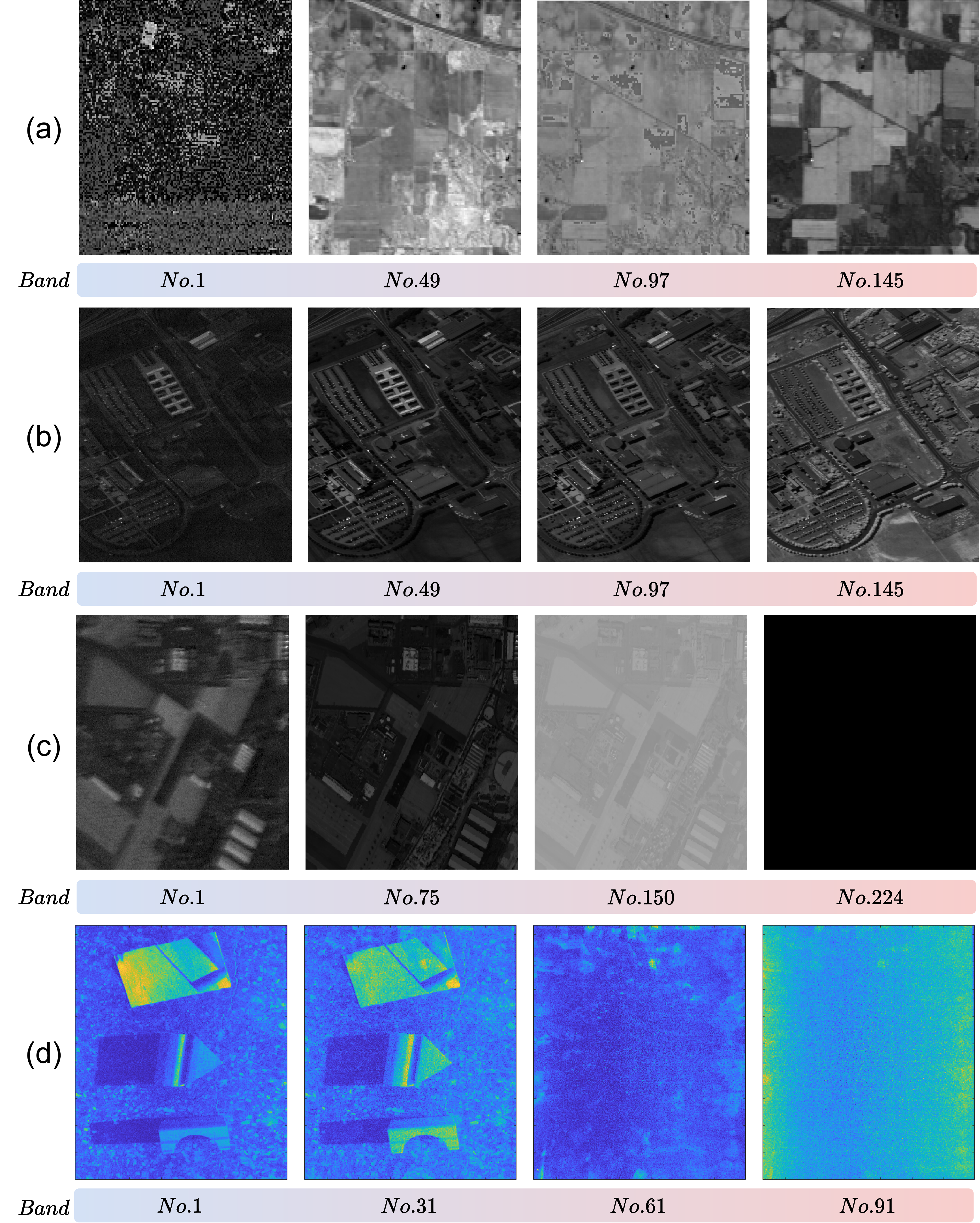}
	\caption {Visualization of differences between bands in four group of images from four corresponding hyperspectral datasets: (a) Indian Pian (b) Pavia University (c) Sandiego (d) HOD-1 dataset. We select a portion of the bands in a fixed ratio to display, covering the visible wavelength range to the infrared wavelength range. The difference in band information increases as the band distance increase.}
	\label{img:dshow}
\end{figure}

The aforementioned methods primarily leverage convolution or attention mechanisms to interact between spectral and spatial information~\cite{8762162}. However, the different bands of hyperspectral images covering varying wavelength ranges result in high heterogeneity of spatial information, especially between bands having long distances. As illustrated in Fig.~\ref{img:dshow}, the different spectral wavelength domains exhibit varying levels of spatial informativeness. Therefore, the inter-band heterogeneity confounds the representation of hyperspectral spatial information. The infrared band (greater than 760 nm) in hyperspectral images primarily reveals heat sources, temperature gradients, and infrared properties of objects when analyzing the temperature and thermal pattern changes~\cite{amenabar2017hyperspectral}. In contrast, the visible band (400-760 nm) provides information about surface features, shapes, and contours with the fine details and textures of objects~\cite{jiang2022feasibility}.

From the spatial perspective, each band of a hyperspectral image represents information in a different wavelength range, and one band corresponds to one modality which is demonstrated in Fig.~\ref{img:dshow}. Besides, according to the characteristics of hyperspectral images, the heterogeneity between bands is more prominent as the distance between bands increases. Hence, the inconsistent semantics of inter-band spatial information limits the performance of object detection in hyperspectral images.

Inspired by the above discussions, we introduce a novel framework named Spectral Discrepancy and Cross-Modal semantic consistency learning (SDCM) to explore an effective and robust band-wise method for hyperspectral images. Specifically, representative bands of infrared and visible light are selected from the hyperspectral images using a band selector and then they are fed into two encoders. After feature encoding, the dynamic cross-modal spatial attention mechanism in the Semantic Consistency Learning (SCL) module is designed to mitigate the inter-band heterogeneity in the hyperspectral images and ensure a unified information representation. On the other hand, the Spectral Gated Generator (SGG) module is integrated to filter the redundant data in hyperspectral inter-band, thus refining the presentation of semantic information.

Furthermore, the complexities in hyperspectral data, encompassing high-dimensional spectral information and intricate background interference, pose considerable challenges for achieving precise object detection~\cite{yue2022spectral, liu2022siamhyper}. To facilitate this, we leverage spectral information to refine the semantic representation of high-level spatial information through a dedicated Spectral Discrepancy Aware (SDA) module. Hence, the SDCM thoroughly accounts for the spectral and spatial attributes of hyperspectral images. Comprehensive experimentation carried out on the HOD-1 and HOD3K hyperspectral datasets validates the superiority of our proposed method over other state-of-the-art methods and demonstrates remarkable performance.

The main contributions can be summarized as follows:

\begin{enumerate}
	\itemsep=0pt
	\item We propose a SDCM framework for object detection in hyperspectral images, comprising SCL, SGG, and SDA modules that synergize together to mine both spatial and spectral information.
 
	\item A SCL module is proposed to mitigate the heterogeneity of inter-bands in hyperspectral images, and then embed the SGG module to filter out the redundant intra-bands information.
 
	\item A SDA module utilizes a priori spectral information from hyperspectral images to correct the semantics of high-level spatial information.
 
	\item Extensive experiments on two public datasets show that the proposed method can mitigate spatial heterogeneity of hyperspectral images, accurately localize targets based on hyperspectral a priori spectral information, and achieve state-of-the-art performance.
\end{enumerate}

The rest of this paper is organized as follows. Section~\ref{sec:2} introduces several typical hyperspectral detections and object detection methods, including one-stage, two-stage, and transformer-based methods. Section \ref{sec:3} describes the details of the proposed method. In Section \ref{sec:4}, experiments on two public datasets and the corresponding results and analysis are presented. Finally, the conclusion is given in Section \ref{sec:5}.

\section{Related work}
\label{sec:2}

\subsection {Generic Object Detection}

Object detection plays a crucial role in extracting valuable information from specific regions of interest~\cite{yan2021object,liu2021anchor}. In the past decade, there have been many excellent object detection works that have been widely applied~\cite{sun2020scalability,koz2019ground}. Specifically, prevailing object detection methods can be broadly categorized into two main groups: one-stage and two-stage detectors~\cite{wang2022remote}. Two-stage detectors comprise two phases, involving the generation of candidate frames followed by the accurate classification and localization of these candidates. Faster R-CNN, introduced by Ke et al.\cite{ren2015faster}, utilizes a region proposal network to create candidate object frames, subsequently refining these frames through bounding box regression to achieve object detection~\cite{zhao2019object}. Conversely, one-stage detectors are well-known for their speed and can be further classified into anchor-free and anchor-based detectors~\cite{padilla2020survey}. YOLO~\cite{redmon2017yolo9000} surpasses conventional two-stage detectors in terms of both speed and accuracy by integrating high-level and low-level semantic information through a mixer module. CenterNet~\cite{duan2019centernet} approaches object detection by focusing on identifying the centroid of the object within the image without relying on anchors~\cite{dai2021up}. Visual transformers, in comparison to CNNs, also exhibit competitiveness within object detection~\cite{guo2022cmt,padilla2020survey,roy2020attention}. Despite the significant success achieved by CNN-based detectors in object detection, their convolutional design introduces inherent biases and certain limitations.

Recently, DETR~\cite{carion2020end} introduced an end-to-end detector framework that yielded extraordinary results and influenced a range of transformer-based detectors. However, the comparatively slower training and inference speed of visual transformers has prompted researchers to explore techniques such as model restructuring to enhance their applicability across a wider array of applications~\cite{zhao2022scene}.To minimize the calculation, Zhu et al.~\cite{zhu2020deformable} proposed deformable attention to accelerate transformer-based detectors. However, despite the central role of object detection within computer vision, its application to hyperspectral imagery remains relatively limited~\cite{ge2022mmsrc}. To this end, we aim to leverage the strengths of transformer and CNN methodologies to develop a hyperspectral object detector.

\subsection{Hyperspectral images based object recognition}

Leveraging the wealth of information offered by numerous continuous spectral bands, hyperspectral imagery holds significant potential for information characterization and interpretation~\cite{9904945,7565539}. However, due to the limitations of hyperspectral image acquisition and resolution, there are few datasets and methods for object detection in hyperspectral data, and existing studies have focused on pixel-level analysis~\cite{fang2023hyperspectral}. Hyperspectral techniques enable the acquisition of spectral information across multiple continuous wavelength ranges, often characterized by high correlation and redundancy among adjacent bands~\cite{sun2022novel}. Existing methods~\cite{ahmad2021hyperspectral} mainly utilize the complementary information of spectral and spatial dimensions based on the multi-band property of hyperspectral images.
Zhang et al.\cite{zhang2023matnet} aims to reduce redundant information in hyperspectral data, facilitating the extraction of information that supports various applications~\cite{liu2019review}.

\begin{figure}
	\centering
	\includegraphics[width = 1\linewidth]{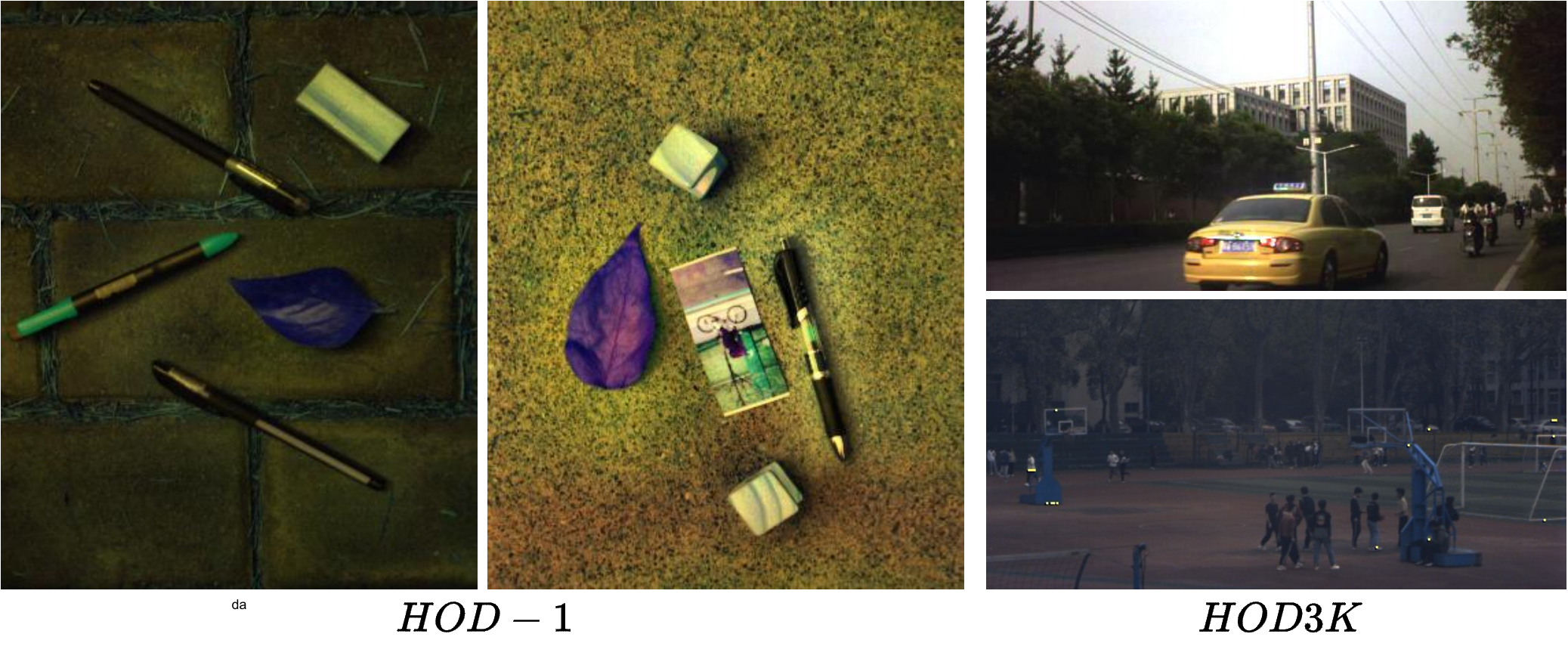}
	\caption {Visualization of sample images in HOD-1 and HOD3K datasets. To better show the hyperspectral images, we transformed the hyperspectral images into pseudo-color images for display.}
	\label{img:dataset_show}
\end{figure}

In the domain of hyperspectral image classification, the pioneering study by Li et al.\cite{chen2014deep} applied the concept of deep learning to hyperspectral images, garnering widespread attention. As deep learning advances in the hyperspectral, the focus of hyperspectral classification methods shifted to the integration of 3D CNNs, 2D CNNs, and self-attention mechanisms. Fang et al.\cite{fang2023hyperspectral} introduced a hyperspectral instance segmentation approach for the semantic segmentation of detected objects. They enhanced network model performance through spectral and spatial attention, alongside the publication of a hyperspectral instance segmentation dataset tailored for remote sensing scenes. Kumar et al.\cite{kumar2023methanemapper} proposed an end-to-end transformer network guided by a spectral linear filter to detect methane gas using the rich spectral information of hyperspectral images. This work alleviates the challenge of greenhouse gas detection. Yan et al.\cite{yan2021object} developed a hyperspectral object detector using a 3D convolution kernel to mitigate overfitting issues stemming from excessive parameters in the hyperspectral feature extraction network. Additionally, they released the HOD-1 dataset, designed for camouflage object detection in hyperspectral images. Recently, Xiao et al.\cite{he2023object} proposed the HOD3K dataset and spectral-spatial joint aggregation method to promote the development of the hyperspectral field further. Limited by the sensor and memory constraints of hyperspectral, HOD3K and HOD-1 are the two existing hyperspectral object detection datasets covering different scenarios, as shown in Fig.~\ref{img:dataset_show}.

However, the aforementioned methods do not take into account the heterogeneity of inter-bands in hyperspectral images. This heterogeneity is exhibited by the difference in spatial information between different bands, which leads to conflicting information between different bands. This limitation hinders the improvement of object detection performance.
\begin{figure*}
	\centering
	\includegraphics[width = 1\linewidth]{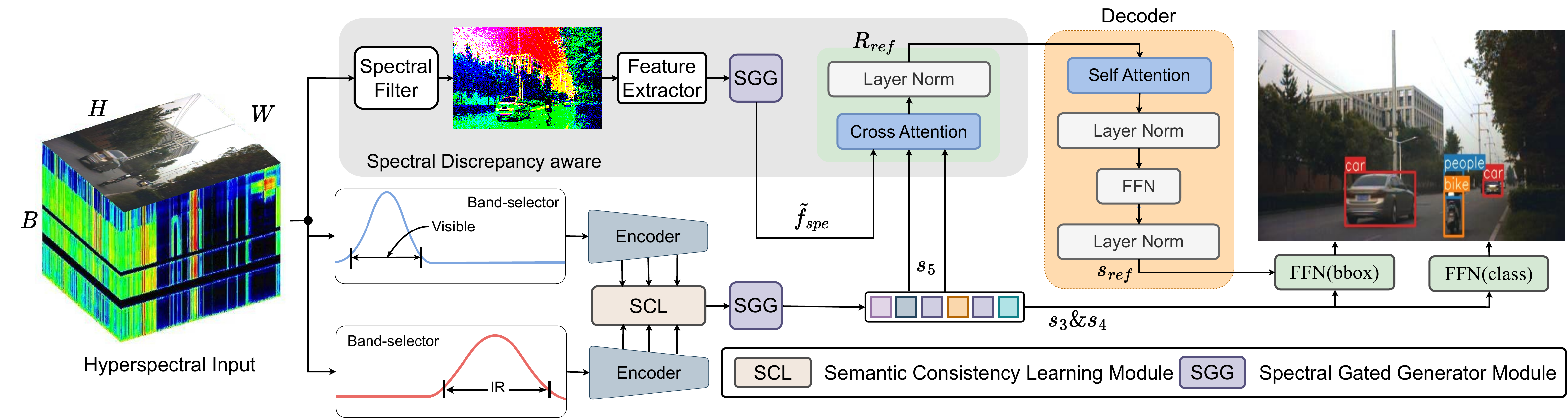}
	\caption {An Overview of Spectral Discrepancy and Cross-modal semantic consistency learning (SDCM). Given a hyperspectral image, we filter out representative visible and infrared (IR) channels by band selection filters and provide the SCL module to mitigate inter-band heterogeneity. The spectral discrepancy aware module accepts input images from all channels and generates spectral information. Next, the spectral information is categorically characterized by cross-attentive guidance of high-level ($\boldsymbol{s}_5$) semantic information using the Transformer decoder. The SGG module is embedded after feature extraction to eliminate redundant information. Lastly, semantically consistent low-level features and refined high-level features are combined and predicted by FFN (feed-forward network) for bounding boxes and category projections.}
	\label{img:backbone}
\end{figure*}
\section{Proposed method}\label{sec:3}
In this section, the overview of the proposed SDCM framework is first introduced. And then, the essential modules in it are described in detail. 

\subsection{Overview}

Fig.~\ref{img:backbone} illustrates the integrated architecture proposed for object detection in hyperspectral images, referred to as the SDCM framework. It mainly comprises five components: two-stream encoders, a spectral discrepancy-aware module, a semantic consistency learning module, a spectral gated generator, and a spectral information decoder. First, the hyperspectral image is passed through a band selection filter~\cite{wang2020hyperspectral} that identifies channels associated with representative visible and infrared bands. Then, the selected visible and infrared bands are fed into a dual-stream encoder separately, producing features in terms of reduced spatial dimensions ($1/2, 1/4, \cdots, 1/32$) denoted as $\{\boldsymbol{s}_i | i=1,2,\cdots,5\}$, where $i$ stands for each stage of feature extraction. To mitigate the problem of spatial information discrepancies between hyperspectral image bands, the outputs of the last three layers of the encoder are fed into the Semantic Consistency Learning (SCL) module which is achieved by a dynamic cross-modal spatial attention mechanism.

After that, the low-level semantic features $\boldsymbol{s}_3$ and $\boldsymbol{s}_4$ are directly fed into the detection head. On the other hand, the SDA module generates comprehensive spectral semantic information by capturing the spectral correlation through the spectral linearity filter, and then this information is cross-attended with the high-level semantic features $\boldsymbol{s}_5$ to refine the high-level semantic representation of spatial information. Lastly, a hyperspectral decoder generates refined high-level semantic features $\boldsymbol{s}_{ref}$, which is forwarded into a feed-forward neural network to predict the location and category of the objects. The SGG module is embedded after the SDA and SCL modules to remove redundant information from the hyperspectral images, thus enhancing the semantic representation of the hyperspectral data after each feature extraction stage.

\begin{figure*}
	\centering
	\includegraphics[width = 1\linewidth]{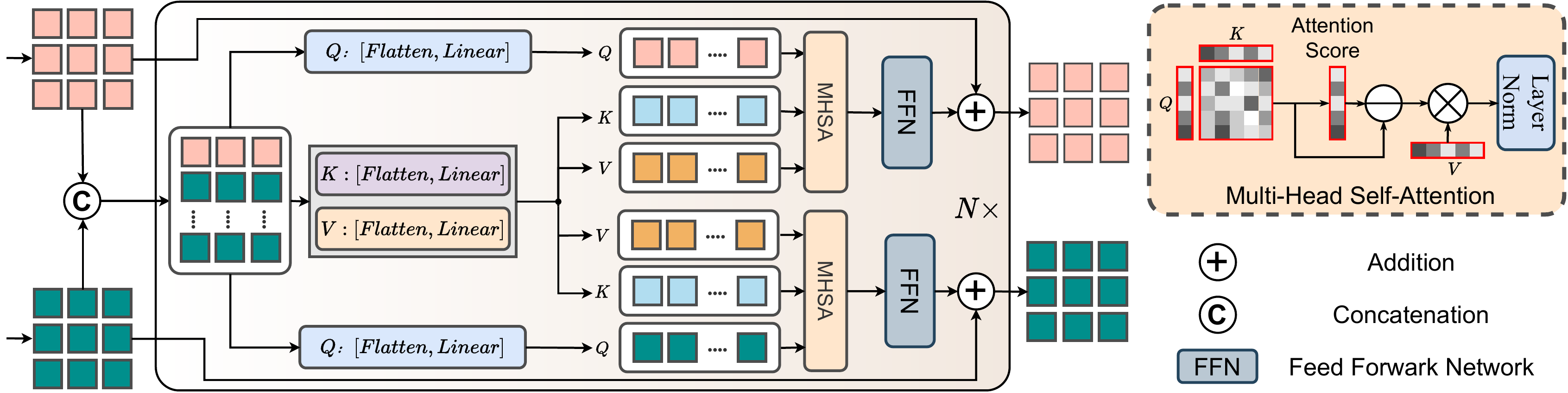}
	\caption {Illustration of the Semantic Consistency Learning (SCL) module.}  	
	\label{img:scl}
\end{figure*}
\subsection{Semantic Consistency Learning Module}
Unlike object detection based on visible images, each band of a hyperspectral image contains specific information, and each band corresponds to a modal of information representation. Therefore, the greater distance between two bands leads to a bigger difference in spatial information. This inter-band spatial inconsistency imposes limitations on object detection in hyperspectral images. In light of this challenge, we introduce the Spatial Consistency Learning (SCL) module, designed to address the heterogeneity in inter-band spatial information.

As depicted in Fig.~\ref{img:scl}, the hyperspectral image undergoes initial processing through two parallel band selection filters~\cite{wang2020hyperspectral} to derive the visible feature $\boldsymbol{s}_{vi} \in \mathbb{R}^{h \times w \times c}$ and the infrared feature $\boldsymbol{s}_{ir} \in \mathbb{R}^{h \times w \times c}$ for the respective visible and infrared band filters. These feature outputs are concatenated to form $\boldsymbol{f}=\textrm{Concat}(\boldsymbol{s}_{vi},\boldsymbol{s}_{ir})$, where $\boldsymbol{f} \in \mathbb{R}^{h \times w \times 2c}$ represents the inter-band consensus information extracted from the hyperspectral data. To mitigate the heterogeneity between visible and infrared features, we employ cross-modal spatial attention, treating $\boldsymbol{s}_{vi}$ and $\boldsymbol{s}_{ir}$ as queries, and $\boldsymbol{f}$ as keys and values. The operations for $\boldsymbol{s}_{vi}$ and $\boldsymbol{s}_{ir}$ are identical and can be formulated as follows:
\begin{equation}
	\begin{gathered}
		\boldsymbol{s}_{vi}^*=\boldsymbol{s}_{vi}^n+\textrm{CMAtt}\left(\textrm{LN}\left(\boldsymbol{s}_{vi}^n\right)\right), \\
    \boldsymbol{s}_{ir}^*=\boldsymbol{s}_{ir}^n+\textrm{CMAtt}\left(\textrm{LN}\left(\boldsymbol{s}_{ir}^n\right)\right),
	\end{gathered}
\end{equation}

where $\textrm{CMAtt}(\cdot)$ represents cross-modal attention, and $\textrm{LN}$ denotes LayerNorm, and the $\boldsymbol{s}_{vi}$ and $\boldsymbol{s}_{ir}$ dual branches multi-head self-attention is sharing weights, to reduce computational redundancy. Subsequently, a nonlinear transformation is applied using the feed-forward network (FFN) to map the features back to their respective waveform representations:
\begin{equation}
	\begin{gathered}
		\boldsymbol{s}_{vi}^{n+1}=\boldsymbol{s}_{vi}^*+\textrm{FFN}\left(\textrm{LN}\left(\boldsymbol{s}_{vi}^*\right)\right),\\
		\boldsymbol{s}_{ir}^{n+1}=\boldsymbol{s}_{ir}^*+\textrm{FFN}\left(\textrm{LN}\left(\boldsymbol{s}_{ir}^*\right)\right),
	\end{gathered}
\end{equation}
where $n \in {1, 2, \cdots, N}$ represents the number of attention stages. This hierarchical interaction through cross-modal spatial attention gradually achieves spatial consistency across hyperspectral bands.

The design of cross-modal attention ($\textrm{CMAtt}(\cdot)$) is carefully tailored to account for the presence of noise interference in hyperspectral data. To ascertain feature importance, we evaluate the similarity between features using matrices $W_Q$ and $W_K$, yielding an attention similarity map denoted as $\boldsymbol{A}$. The principal steps of cross-modal attention are outlined as follows:
\begin{equation}
	\boldsymbol{A} = \boldsymbol{s}^n{ W_Q } \cdot \left( \boldsymbol{s}^n{ W_K } \right)^{ \top }.
 \end{equation}
 The attention matrix is equivalent to the similarity graph and contains the pixel node distance metric, and then we select the top-k elements row-wise of $\boldsymbol{A}$ to compute to obtain the affinity matrix $\boldsymbol{S}$. The process can be formulated as:
 \begin{equation}
\left[\boldsymbol{S}\right]_{i j}= \begin{cases}1, & \text {if}~ \boldsymbol{A}_{i j} \in \text { top-k }  \\ -\infty & \text { otherwise. }\end{cases}
\end{equation}
where $\boldsymbol{S}$ signifies the attention affinity score, initialized with negative infinity. The aim is to mine spatial and spectral coherence information by aggregating attention coefficients based on the hyperspectral attention weight matrix. Subsequently, we selectively retain features rich in semantic information and combine them with values to derive attention coefficients:
\begin{equation}
\textrm{CMAtt}\left(\boldsymbol{f}\right)=\textrm{Softmax}\left(\frac{ { \boldsymbol{A} \cdot \boldsymbol{S} }}{\sqrt{d}}\right) \cdot \boldsymbol{f}{W_V},
\end{equation}
the inclusion of the scaling factor $1 / \sqrt{d}$ serves to adjust the scaling of attention similarity. Compared to standard transformer modules, these modules reduce computational complexity $12 \hat{n} c^2+2 \hat{n}^2 c$ to $10 \hat{n} c^2\left(1+0.2 / k^2\right)+1.75 \hat{n}^2 c / k^2$ while maintaining accuracy, where k denotes the ratio of top-k in tokens. By filtering out inconsequential features through a weighted fraction, the impact of hyperspectral noise information in spatial attention calculation is effectively alleviated.

\begin{figure}
	\centering
	\includegraphics[width = 0.86\linewidth]{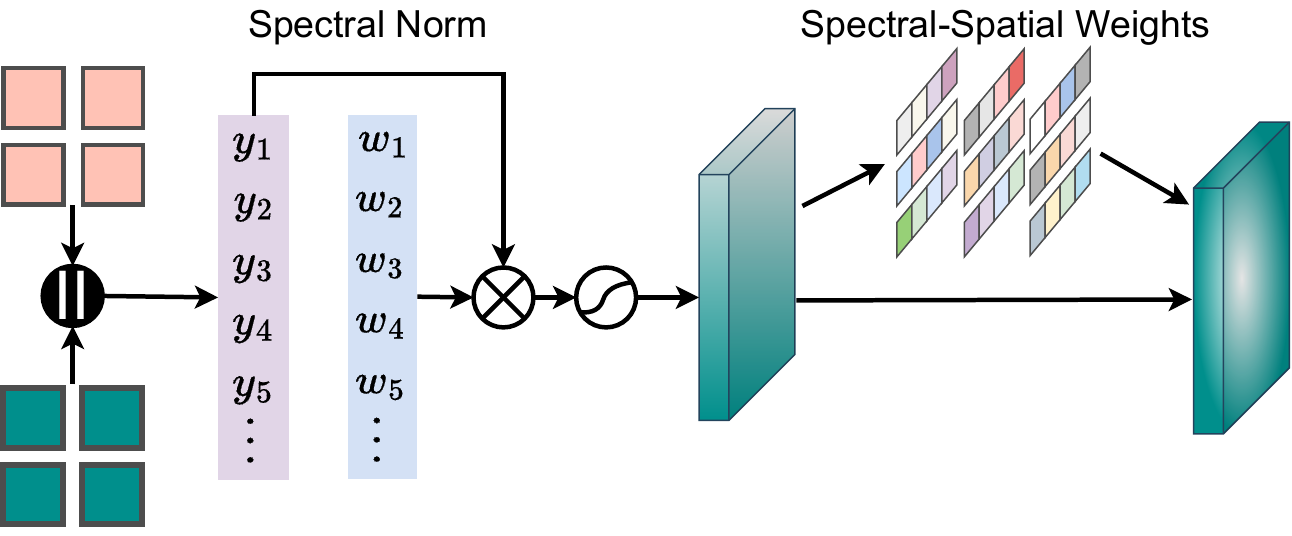}
	\caption {Illustration of the Spectral Gated Generator (SGG) module. The weights are obtained based on the assumption that all pixels in a single band follow the same distribution.}	
	\label{img:sgg}
\end{figure}
\subsection{Spectral Gated Generator Module}
Hyperspectral bands are typically densely packed, with a narrow interval of approximately five nanometers separating them~\cite{shukla2016overview}. As a result of this close spacing, the spectral characteristics of various targets often converge within the same pixel. Therefore, neighboring bands exhibit a notable level of redundant information, giving rise to substantial similarities and duplication of spectral attributes within hyperspectral images. To mitigate this issue, we introduced the Spectral Gated Generator (SGG) module to assess the significance of spectral information and eliminate redundant spectral content, as illustrated in Fig.~\ref{img:sgg}. First, we concatenate the features $\boldsymbol{s}_{vi}$ and $\boldsymbol{s}_{ir}$ that have passed through the SCL module, denoting the concatenated result as $\boldsymbol{s}$. This process is mathematically expressed as follows:
\begin{equation}
	\widetilde{\boldsymbol{s}} = \textrm{Sigmoid}\left(W_\gamma^i\left(\textrm{BN}(\boldsymbol{s})\right)\right),
\end{equation}
where $\textrm{BN}(\cdot)$ denotes the BatchNorm function, and the weights $W_\gamma$ are set according to $W_\gamma = \gamma_i / \sum_{j=0} \gamma_j$. Inspired by~\cite{yang2021simam}, we adopt a parameter-free 3D attention mechanism to facilitate interaction between spatial and spectral information. The energy function is defined as:
\begin{equation}
	e_t^* = \frac{4\left(\hat{\sigma}^2+\lambda\right)}{(t-\hat{\mu})^2+2 \hat{\sigma}^2+2 \lambda},
\end{equation}
where $\hat{\mu}=\frac{1}{M} \sum_{i=1}^M x_i$ and $\hat{\sigma}^2=\frac{1}{M} \sum_{i=1}^M\left(x_i-\hat{\mu}\right)^2$. 

This equation reveals that a lower energy value $e_t^*$ indicates a more pronounced distinction between neuron $t$ and its neighboring neurons, underscoring its significance in visual processing. Consequently, the importance of each neuron can be inferred from $1 / e_t^*$. To further enhance the process, a scaling operator is introduced:
\begin{equation}
	{\boldsymbol{s}_t} = \textrm{Sigmoid}\left(\frac{1}{\mathbf{E}}\right) \odot \widetilde{\boldsymbol{s}},
\end{equation}
where $\mathbf{E} = e_1,e_2,\cdots,e_t,$ represents the collection of all $e_t^*$ values across the spectral and spatial dimensions.

\begin{table*}[t]
	\centering
	\caption{  Comparison to methods on HOD-1 dataset. The screen is denoted as S, and real is denoted as R. TL represents transfer learning, where hyperspectral cubes for migration learning are generated and pre-trained by a channel-wise expansion of RGB images and then fine-tuned on real hyperspectral datasets, enabling the model to learn rich spatial domain information from a large number of RGB images.}
	\begin{tabular}{c|c|cccccccccc} 
		\toprule[1.5pt]
		Detectors&Backbone &toyblock S&photo S&pen S&photo R&toyblock R&pen R&leaf S&leaf R & mAP@.5 \\
		\midrule
       {Retinanet~\cite{lin2017focal}} & VGG16&42.4& 58.9& 24.8 &43.4&83.3 &49.1& 31.0& 80.0& 51.6\\
		{FCOS~\cite{tian2019fcos}} & ResNet50&-&-&-&-&-&-&-&-&{80.9} \\
  	{Deformable DETR~\cite{zhu2020deformable}} & VGG16&71.4&70.4&56.2&87.4&80.6&83.8&46.8&76.6&71.7\\
   
        {Faster RCNN~\cite{ren2015faster}} &ResNet50&72.2&80.0&44.7&95.1&92.4&80.2&72.8&93.1&78.8\\
 		{Double-Head RCNN~\cite{wu2020rethinking}} &ResNet50&-&-&-&-&-&-&-&- & {81.2} \\

      {Cascade RCNN~\cite{cai2019cascade}}& ResNet50&89.6&78.6&55.1&94.9&94.0&88.3&80.8&93.3&84.3 \\
      
  	{YOLOF~\cite{chen2021you}} & DarkNet53&87.4&75.0&39.7&96.4&90.6&78.2&82.3&92.6&80.3 \\
   
       {Libra RCNN~\cite{pang2019libra}} &ResNet50&90.3& 91.8& 62.4&96.9&95.3&79.3&85.5&95.5&87.1 \\
	{FoveaBox~\cite{kong2020foveabox}} & ResNet50&-&-&-&-&-&-&-&- & {80.2} \\

	{YOLOv5} & DarkNet53 &82.4&80.3&34.0&97.7&97.6&76.9&81.1&93.3&80.4 \\
		{HOD-1~\cite{yan2021object}} & VGG16&-&-&-&-&-&-&-&- & {83.5} \\

		{S2ADet~\cite{he2023object}} & DarkNet53 &{83.1}&{88.3}&{51.5}&\textbf{98.4}&\textbf{98.0}&{91.2}&{83.3}&\textbf{98.7}&{86.6} \\	
    	{MethaneMapper~\cite{kumar2023methanemapper}} & ResNet50&82.1&90.4&60.9&89.4&87.5&90.6&86.1&89.1&84.5\\
						\midrule
		{HOD-1 (TL)~\cite{yan2021object}} & VGG16&-&-&-&-&-&-&-&- & {89.3} \\
		\midrule
		{SDCM (Ours)} & DarkNet53 &\textbf{93.0}&\textbf{94.8}&\textbf{83.5}&{97.5}&{97.6}&\textbf{94.9}&\textbf{89.0}&{98.4}&\textbf{93.6} \\
		\bottomrule[1.5pt]
	\end{tabular}
	\label{tab:HOD-1}
\end{table*}

\subsection{Spectral Discrepancy Aware Module}

The spectral dimension of hyperspectral imagery yields valuable insights into material properties, facilitating effective object classification. However, conventional network models often prioritize spatial information, neglecting the spectral dimension during feature extraction. The proposed Spectral Discrepancy Aware (SDA) module aims to rectify this limitation by aggregating pixel-level spectral information to guide high-level semantic features, improving the accuracy of object category prediction.

Given the input hyperspectral image $\boldsymbol{f}_{spe} \in \mathbb{R}^{h \times w \times b}$, with $b$ denoting the number of spectral bands, the SDA module initially aggregates spectral information using a spectral filter, such as principal component analysis (PCA)~\cite{wold1987principal}. Then, spectral information features are obtained through a feature extractor, i.e., ResNet~\cite{he2016deep}, leading to the following formulation:
\begin{equation}
\widetilde{\boldsymbol{f}}_{spe}=\textrm{FeatureExtractor}\left(\textrm{Spectral-Filter}\left(\boldsymbol{f}_{spe}\right)\right).
\end{equation}
Next, the SGG module selects semantically informative features based on the importance of the spectral information. To incorporate the spectral information $\widetilde{\boldsymbol{f}}_{spe}$ into high-level feature representation, a cross-focus mechanism is employed. Drawing inspiration from \cite{sun2022spectral,salvador2021revamping}, the design follows a transformer encoder-decoder architecture. The spectral information $\boldsymbol{f}_{spe}$ is utilized as a query, while the high-level feature $\boldsymbol{s}_5$ serves as a key and value. The inputs for the cross-attention operation are $\boldsymbol{f}_{spe}$ and high-level feature ($\boldsymbol{s}_5$), and the computation is defined as:
\begin{equation}
	\textrm{Cross-Att}(\widetilde{\boldsymbol{f}}_{spe},{\boldsymbol{s}_5})=\textrm{softmax}\left(\frac{ {\widetilde{\boldsymbol{f}}_{spe}W_Q \cdot \boldsymbol{s}_5 W_K }}{\sqrt{d}}\right) \boldsymbol{s}_5{W_V}.
\end{equation}
Then, the refined high-level features $R_{ref}$ are obtained and then input to the decoder module. The detector employs a self-attention mechanism to capture dependencies within the object sequence and generate output embedding. This process is represented by the equation:
\begin{equation}
	\boldsymbol{s}_{out}=\textrm{Decoder}\left(R_{ref}, p\right),
\end{equation}
where $p$ represents the positional embedding. Lastly, high-level semantic features undergo correction based on spectral information guidance.

\subsection{Loss Function.}
In the domain  of hyperspectral image object detection, the formulation of the multitask loss function during the training phase is articulated as:
\begin{equation}
\begin{aligned}
    \mathcal{L} &= \mathcal{L}_{\mathrm{cls}} + \mathcal{L}_{\mathrm{box}} + \mathcal{L}_{\mathrm{conf}}, \\
    \mathcal{L}_{\mathrm{cls}} &= -\sum_{i=0}^{N} p_i \log(\hat{p}_i), \\
    \mathcal{L}_{\mathrm{conf}} &= -\sum_{j=0}^{K} c_j \log(\hat{c}_j) + (1 - c_j) \log(1 - \hat{c}_j).\\
    \mathcal{L}_{\mathrm{box}} &= \sum_{i=0}^{M} \mathrm{SmoothL1}(t_i - \hat{t}_i), \\
\end{aligned}
\end{equation}

where $\mathcal{L}_{\mathrm{cls}}$ denotes the classification loss~\cite{lin2017focal}, and $\mathcal{L}_{\mathrm{conf}}$ is the confidence loss. $\mathcal{L}_{\mathrm{box}}$ represents the bounding-box loss~\cite{rezatofighi2019generalized}, where $\mathrm{SmoothL1}(x) = 
\begin{cases} 
0.5 \cdot x^2 & \text{if } |x| < 1, \\
|x| - 0.5 & \text{otherwise}.
\end{cases}$. 
$\mathcal{L}_{\mathrm{conf}}$ stands for loss of confidence~\cite{redmon2016you}, which is commonly used to measure the accuracy of classification. This structured approach, which distinctly addresses bounding-box localization, object classification, and confidence scoring, diverges from traditional object detection methodologies. Previous research has highlighted the critical role of such a loss function in augmenting the performance of object detection models. Our research demonstrates that this loss function migrates to hyperspectral images and works cooperatively to localize the target of interest.

\section{Experiment}
\label{sec:4}
This section presents a comprehensive series of experiments conducted on two datasets to validate the efficacy of the proposed SDCM framework. We commence by introducing several publicly available datasets along with the set of algorithms used for comparison in the experiments. Subsequently, the specifics of the experimental configuration are outlined. Finally, a detailed analysis of the outcomes obtained by various algorithms across distinct datasets is presented, conclusively establishing the superior performance of our proposed method.
\subsection{Datasets}

\subsubsection{HOD-1 Dataset~\cite{yan2021object}} is tailored for camouflage object detection within hyperspectral images. This dataset encompasses 1657 images, systematically classified into eight distinct categories, each represented across a comprehensive spectral range of 96 bands. Construction of this dataset involved the strategic placement of objects to formulate intricate scenes, subsequently captured via a hyperspectral camera. Notably, scenes with camouflage were meticulously devised using a two-step procedure. Initially, a real-world scene was captured utilizing an iPadAir camera, which was projected onto the iPadAir screen and captured anew via the hyperspectral camera. The intrinsic spectral characteristics of hyperspectral images facilitated the differentiation between genuine and camouflaged scenes, showcasing the dataset's robust classification potential.

\subsubsection{HOD3K Dataset~\cite{he2023object}} comprises 16 spectral bands, portraying diverse natural scenarios encompassing roadways, residential zones, and sports fields. A comprehensive collection of 3242 hyperspectral images was curated, meticulously partitioned into 2308 training samples, 219 for validation, and 715 designated for rigorous testing. The imaging process was executed through the employment of the XIMEA snapshot VIS camera. Notably, the dataset encapsulates a total of 15,149 distinct objects, translating to an average of 4.67 objects per image. This trove encompasses 12,144 individuals, 817 vehicles, and 2,188 bicycles, showcasing the dataset's remarkable diversity and utility for object detection studies.


\begin{table*}[t]
	\caption{Comparison to methods on HOD3K dataset with resolution at 512 $\times$ 256. SA denotes the image of spectral aggregated information, and SE denotes the image of spectral aggregated information.}
	\begin{center}
		\begin{tabular}{c|c|c|cccc|c|c|c}
			\toprule[1.5pt]
			Detectors &Backbone&Type  & people &bike & car & mAP@.5 &  mAP@.5:.95   &FLOPs  & Param. 	\\
			\midrule
			{Faster RCNN~\cite{ren2015faster}} &ResNet50 &two-stage& {81.8} & {94.5} &{91.7}&{89.4}&{56.9}&206.68 &41.14M \\ 
			{Libra RCNN~\cite{pang2019libra}} &ResNet50 &two-stage& {83.6} & {95.0} &{90.8}&{89.8}&{56.6} &207.73 &41.40M \\ 
   		{Cascade RCNN~\cite{cai2019cascade}} & ResNet50 &two-stage& {84.5} & {96.1} &{90.6}&{90.4} & 57.4 & 57.4 &42.52M \\
			\midrule
			{FCOS~\cite{tian2019fcos}} & ResNet50&one-stage& {55.0} & {19.7} &{69.8}&{48.2}&{23.7}& 196.81 &31.84M  \\
			
			{YOLOF~\cite{chen2021you}} & ResNet50 &one-stage& {60.8} & {67.8} &{67.6}&{65.4}&28.6 &98.23 &{42.13M}\\	 
		{Deformable DETR~\cite{zhu2020deformable}} & ResNet50 &one-stage& {52.8} & {56.3} &{64.9}&{58.0} &22.3&195.23&{39.82M}\\	

			{Retinanet~\cite{lin2017focal}} &ResNet50&one-stage & {85.6} & {94.8} &{92.6}&{91.2}&{53.3}&205.69 &36.17M\\ 
			{TOOD~\cite{feng2021tood}} &ResNet50 &one-stage& {85.1} & {87.0} &{89.6}&{87.2}&{55.4}&180.66&31.80M\\
			{YOLOv5} & DarkNet53 &one-stage& {79.3} & {94.0} &{91.2}&{88.1}&{54.4} &\textbf{48.30} &\textbf{20.88M}\\	 
			\multicolumn{1}{c|}{YOLOv5\dag} & \multicolumn{1}{c|}{DarkNet53} &\multicolumn{1}{c|}{one-stage}& {83.6} & {96.4} &{95.2}&{91.7}&{56.3}& 89.72 &35.49M \\	
			{S2ADet~\cite{he2023object}} & S2ANet &one-stage& {87.2} & {97.7} &{95.3}&{93.4}&{59.8} &169.20 & 48.64M \\
           {MethaneMapper~\cite{kumar2023methanemapper}}& ResNet50 &one-stage & {80.2} &{95.1}& {90.9}&{88.7} & 57.9 &267.64  & 80.32M   \\
		\midrule
		{SDCM (Ours)} & DarkNet53 &one-stage& \textbf{91.2} & \textbf{98.1} &\textbf{96.0}&\textbf{95.2} &\textbf{60.6}  &181.51  & 53.94M  \\
			\bottomrule[1.5pt]
		\end{tabular}
	\end{center}
	\label{tab:HOD3K}
\end{table*}

\begin{table*}[t]
	\centering
	\caption{Ablation study of the SDA module in on HOD-1 Dataset. The baseline detector without SCL, SGG, and SDA modules, with visible and infrared information obtained from the band selection filter as input. The screen is denoted as S, and real is denoted as R.}
	\begin{tabular}{c|c|c|ccccccccc}
		\toprule[1.5pt]
		Detectors & SCL& SDA &toyblock S&photo S&pen S&photo R&toyblock R&pen R&leaf S&leaf R& mAP@0.5 \\
		\midrule	
		{Baseline} & $-$ & $-$&88.4&93.0&51.3&96.5&96.8&81.4&71.8&94.6&84.2\\ 
		\midrule	
		{\multirow{3}[0]{*}{SDCM}}& $-$& \ding{52} 	&{91.0}&93.2&{71.9}&\textbf{98.1}&\textbf{98.4}&93.0&59.5&{97.1}&87.8\\ 
		{} & \ding{52} & $-$&{89.9}&\textbf{96.3}&{78.6}&{97.8}&{97.2}&\textbf{97.4}&\textbf{91.9}&{95.7}&{93.1} \\
		
		{} & \ding{52} & \ding{52}&\textbf{93.0}&{94.8}&\textbf{83.5}&{97.5}&{97.6}&{94.9}&{89.0}&\textbf{98.4}&\textbf{93.6} \\
		\bottomrule[1.5pt]
	\end{tabular}
	\label{tab:sda}
\end{table*}

\begin{table}[t]
	\centering
	\caption{Ablation study of SDCM on the HOD-1 dataset. }
	\begin{tabular}{c|ccc|c}
		\toprule[1.5pt]
		Detectors &SGG  & SCL & SDA & mAP@0.5 \\
		\midrule	
			\multirow{5}{*}{SDCM}&- & - &- &{84.2} \\ 
	&\ding{52} & - &- &{86.0} \\ 
 	&- & - & \ding{52} &{87.1} \\ 
		{ } & & \ding{52} &\ding{52}  &{92.5} \\
		{}&\ding{52} & \ding{52} & \ding{52}  &\textbf{93.6}\\ 
		\bottomrule[1.5pt]
	\end{tabular}
	\label{tab:sgg}
\end{table}

\begin{table}[t]
	\centering
	\caption{Ablation study of K parameters in the SCL module on the HOD-1 dataset. The number of top-k is obtained by multiplying the token dimension with the percentage.}
	\begin{tabular}{c|c|ccc}
		\toprule[1.5pt]
	Detectors &Percentage of top-k (\%) & mAP@0.5 \\
		\midrule	
		{\multirow{4}[0]{*}{SDCM}}&	{25}&{92.7} \\ 
		&{50} &\textbf{93.6} \\ 
		&{75} &{93.2} \\
		&{100} &{93.0}\\ 
		\bottomrule[1.5pt]
	\end{tabular}
	\label{tab:k}
\end{table}

\subsection{Experimental Setup}

We adopted a two-stream encoder, initializing it with the pre-trained DarkNet-53 weights~\cite{redmon2018yolov3}. Our encoder has five layers, and the input image size is set to 640 $\times$ 640. The design of the detector and FPN takes into account conventional object detection schemes. The SCL module has a K setting of 100, and the number of modules is set to 2, refer to~\cite{wang2022kvt}. To improve the network performance, we employ data mosaic as a data augmentation technique during the training process. The batch size was set to 8 and the network was trained for 50 epochs with the learning rate of 0.01, which was later adjusted to 0.03. Our comprehensive experimentation was conducted on two distinct hyperspectral datasets, serving to showcase the efficacy of our proposed approach. To experimental consistency, the input image resolution aligns with~\cite{he2023object}. Employing a diverse mosaic of data augmentation techniques, we ensured a fair basis for comparison.

All implementations in this study were grounded in PyTorch 1.12.1 and Python 3.8. The experimentation was carried out utilizing an NVIDIA GeForce RTX 3090 GPU, employing an SGD optimizer with an initial learning rate of 0.01. We adopted the $poly$ learning scheme, dynamically adjusting the learning rate as $(1 - \frac{epoch}{max_epoch})^{power} \times lr$, where $power = 0.9$, and the learning rate was set to 0.05. During the testing phase, a non-maximum suppression intersection over union threshold of 0.6 was applied.

\textbf{Evaluation Metrics:} The assessment metrics are drawn from COCO-Style~\cite{lin2014microsoft}, encompassing the average accuracy across various IoU thresholds, such as mAP@IoU=0.5 and mAP@IoU=0.5:0.95. Furthermore, a comprehensive analysis of the classification confusion matrix was performed to further evaluate the model's classification capability.

\subsection{Comparison With State-of-the-Art}

We present the experimental results of various object detection network models on the HOD-1 and HOD3K datasets, to illustrate the efficacy of our proposed Spectral Discrepancy and Cross-modal semantic consistency aware network (SDCM) in addressing hyperspectral band heterogeneity and integrating spectral and spatial information for enhanced detection performance. The outcomes of our experiments on the HOD-1 and HOD3K datasets are presented in Table~\ref{tab:HOD-1} and Table~\ref{tab:HOD3K}.

To validate the effectiveness of our proposed method, we conducted a comparative analysis against 13 contemporary state-of-the-art methods, including both object detection and hyperspectral object detection. That is YOLOv5, Faster RCNN~\cite{ren2015faster}, Libra RCNN \cite{pang2019libra}, FCOS \cite{tian2019fcos}, Retinanet \cite{lin2017focal}, TOOD \cite{feng2021tood}, Cascade RCNN \cite{cai2019cascade}, Deformable DETR~\cite{zhu2020deformable}, YOLOF \cite{chen2021you}, Double-Head RCNN~\cite{wu2020rethinking}, FoveaBox~\cite{kong2020foveabox}, HOD-1~\cite{yan2021object}, and S2ADet~\cite{he2023object}. For both comparative and qualitative analyses, all data utilized were sourced from resources and code published by the respective authors.

\subsubsection{HOD-1}
As demonstrated in Table~\ref{tab:HOD-1}, our method significantly surpasses other approaches, exhibiting a noteworthy 10.0\% mAP@.5 improvement over the leading method (S2ADet~\cite{he2023object}), thereby indicating the superiority of our approach. Our method attains a state-of-the-art performance with a mAP@.5 of 93.6\%. Notably, our algorithm excels in 5 categories, particularly demonstrating exceptional capability in detecting samples with low metrics. Remarkably, the pen screen category exhibits a substantial 32.0\% mAP@.5 improvement, while the toyblock screen, photo screen, pen real, and leaf screen categories have improvements of 9.9\%, 6.5\%, 3.7\%, and 85.7\% mAP@.5 respectively.

These outcomes underscore the effectiveness of the Spectral Discrepancy Aware module in enhancing the high-level features of hyperspectral images for object classification, resulting in outstanding classification performance. Furthermore, semantic cross-modal consistency learning contributes to object detection and classification by mitigating hyperspectral semantic expression conflicts, thus synergizing with the SDA module.

Notably, our method achieves a 3.8\% mAP@.5 improvement over the HOD-1 (TL) method, which leverages transfer learning from RGB images to hyperspectral images. This highlights the distinct data distributions between hyperspectral and visible data, suggesting that conventional visible image processing techniques may not be directly applicable to hyperspectral detection scenarios. The superior performance of our method in HOD-1 emphasizes its effectiveness, establishing it as a cutting-edge detection solution.

\begin{figure*}
	\centering
	\includegraphics[width = 1\linewidth]{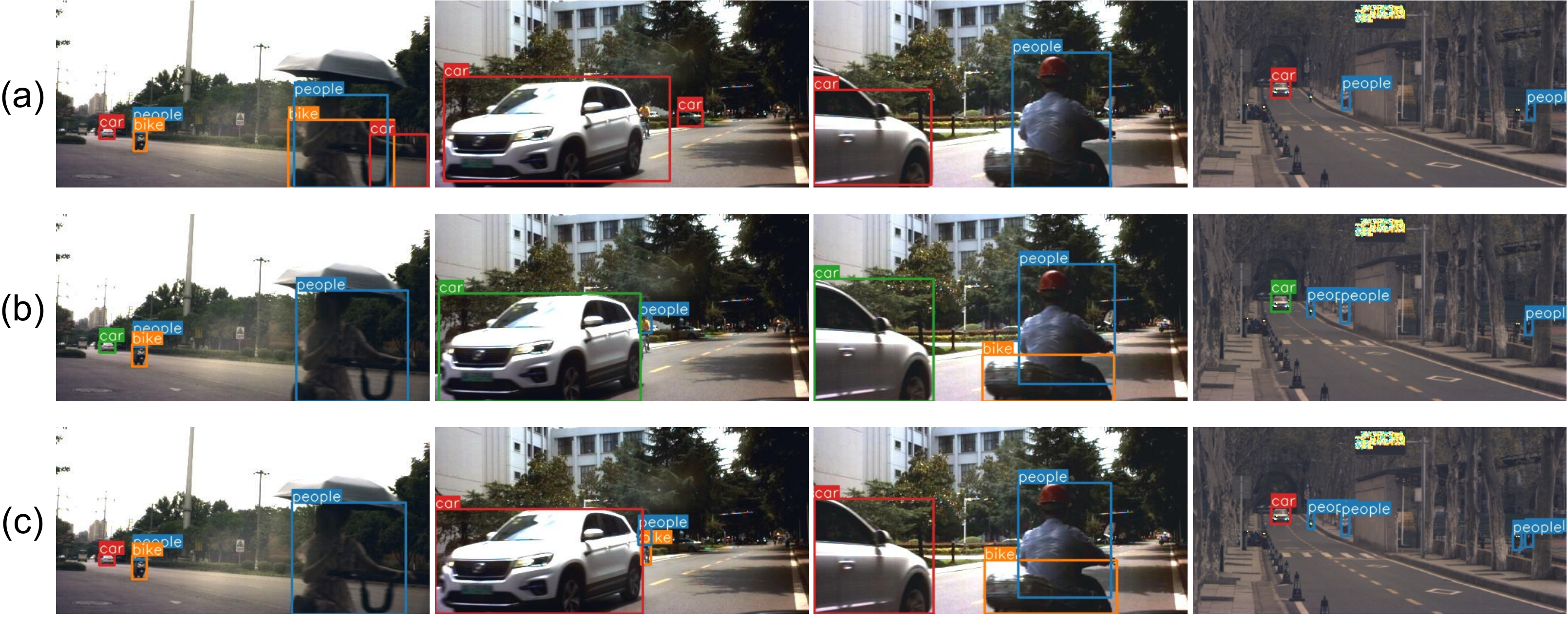}
	\caption {Comparison of the visualization results of different methods on the HOD3K (a) Baseline (b) S2ADet (c) SDCM. The three categories of people, bike, and car are marked with blue, orange, and red (S2ADet is green) object boxes respectively. To better present the detection results, we transformed the hyperspectral images into pseudo-color images for display.} 	
	\label{img:contrast}
\end{figure*}
\begin{figure}
	\centering
	\includegraphics[width = 0.90\linewidth]{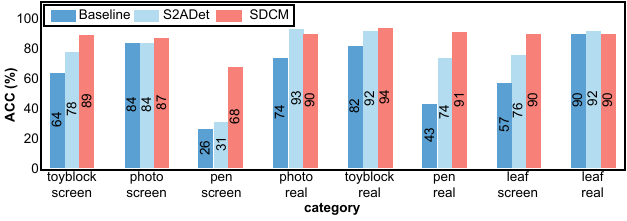}
	\caption {Comparison of the classification accuracy of different methods on the HOD-1 dataset. The HOD-1 is a camouflaged object detection dataset with small gaps between categories, and determining the authenticity of objects is a big challenge.}	
	\label{img:class}
\end{figure}

\subsubsection{HOD3K}

As illustrated in Table~\ref{tab:HOD3K}, all variations of our proposed methods outperform the existing state-of-the-art approaches under identical experimental conditions. Compared to the state-of-the-art algorithm S2ADet, our method boasts approximately 4.35 million additional parameters, and equivalent to 8.9\% of the total model size, yet outperforms it by approximately 1\% mAP@.5:.95 in terms of performance.

In contrast to the two-stage detector, SDCM achieves a noteworthy enhancement in detection accuracy, surpassing the two-stage algorithm in all categories. Moreover, SDCM exhibits a substantial 38.3\% mAP@.5:.95 improvement over the transformers-based detector, i.e., Deformable DETR~\cite{zhu2020deformable}, highlighting the synergistic integration of convolution and transformers. This combination accelerates the convergence of transformers-based detectors, effectively harnessing the dynamic global modeling capabilities of transformers to extract spatial and spectral complementary information from the hyperspectral images. Furthermore, our proposed method attains the highest performance in each category, achieving enhancements of 4.0\%, 0.4\%, and 0.7\% mAP@50 in the people, bike, and car categories, respectively. These results underscore the robustness and efficacy of our module across diverse categories.

\subsection{Ablation Study}

To validate the effectiveness of the proposed SDCM components and configurations, we conducted a comprehensive ablation experiment using the HOD-1 dataset.

\subsubsection{Effectiveness of Semantic Consistency Learning Module (SCL)}

We conducted ablation experiments on the SCL module, as depicted in Table~\ref{tab:sda}. Initially, we integrated the SCL module into the Baseline model. A comparison between the first and third rows in Table~\ref{tab:sda} highlights the contribution of the SCL module, resulting in a significant 8.9\% mAP@.5 improvement. By assimilating hyperspectral information through consistent learning, the SCL module facilitates consistent information representation, thereby mitigating heterogeneity across hyperspectral bands. This proves especially advantageous for precise object boundary localization in low-level features, leading to superior performance across all categories compared to the Baseline.

Additionally, the SCL module alone improves performance by 5.3\% mAP@0.5 compared to using only the SDA module, highlighting the critical role of continuous hyperspectral image learning in object detection. Furthermore, SDCM achieves an additional 0.5\% mAP@0.5 improvement when combined with the SCL module, demonstrating its synergy with object classification and localization in SDCM.

Table~\ref{tab:k} presents the ablation study results, evaluating the impact of varying top-k percentages on the SCL module's performance on the HOD-1 dataset, measured by the mAP@0.5 metric. The top-k value is determined by multiplying the token dimension by the specified percentage, representing the proportion of tokens used in the detection process. The analysis reveals that performance improves as the top-k percentage increases up to 50\%, achieving the highest mAP@0.5 score of 93.6\%. However, beyond this threshold, performance begins to decline slightly, with mAP@0.5 values of 93.2\% and 93.0\% at top-k percentages of 75\% and 100\%, respectively. This suggests that while increasing the number of tokens initially benefits the model, using too many tokens may introduce noise or redundant information, causing a slight reduction in accuracy. Hence, optimal performance is achieved with a top-k percentage of 50\%, underscoring the importance of balancing token selection to maximize detection accuracy while minimizing the inclusion of irrelevant hyperspectral data.

\subsubsection{Effectiveness of Spectral Discrepancy Aware (SDA) Module }

To validate the ability of the SDA module to capture spectral information within hyperspectral images, we conducted experiments both with and without the SCL module. As evident from the first and second rows of Table~\ref{tab:sda}, the SDA module excels in six categories, particularly in the photo real and toyblock real categories, surpassing the full SDCM. This affirms the SDA module's efficient utilization of spectral information to aid high-level features in object categorization. Furthermore, when combined with the SCL module, the SDA module enhances the learning of spectral dimensions in hyperspectral images, thereby achieving even better performance.

\subsubsection{Effectiveness of Spectral Gated Generator (SGG)}

To verify the capacity of the SGG module in processing redundant hyperspectral information, we introduced and removed SGG modules in the Baseline and SDCM, respectively, as demonstrated in Table~\ref{tab:sgg}. Remarkably, even without architectural design changes, Baseline+SGG yields a 1.8\% mAP@.5 improvement over the baseline. This versatility suggests that the SGG module can be seamlessly integrated into various hyperspectral image processing applications. Additionally, SDCM outperforms SDCM without SGG module by 1.1\% mAP@.5, emphasizing the efficacy of combining SGG module.

\begin{figure*}
	\centering
	\includegraphics[width = 1\linewidth]{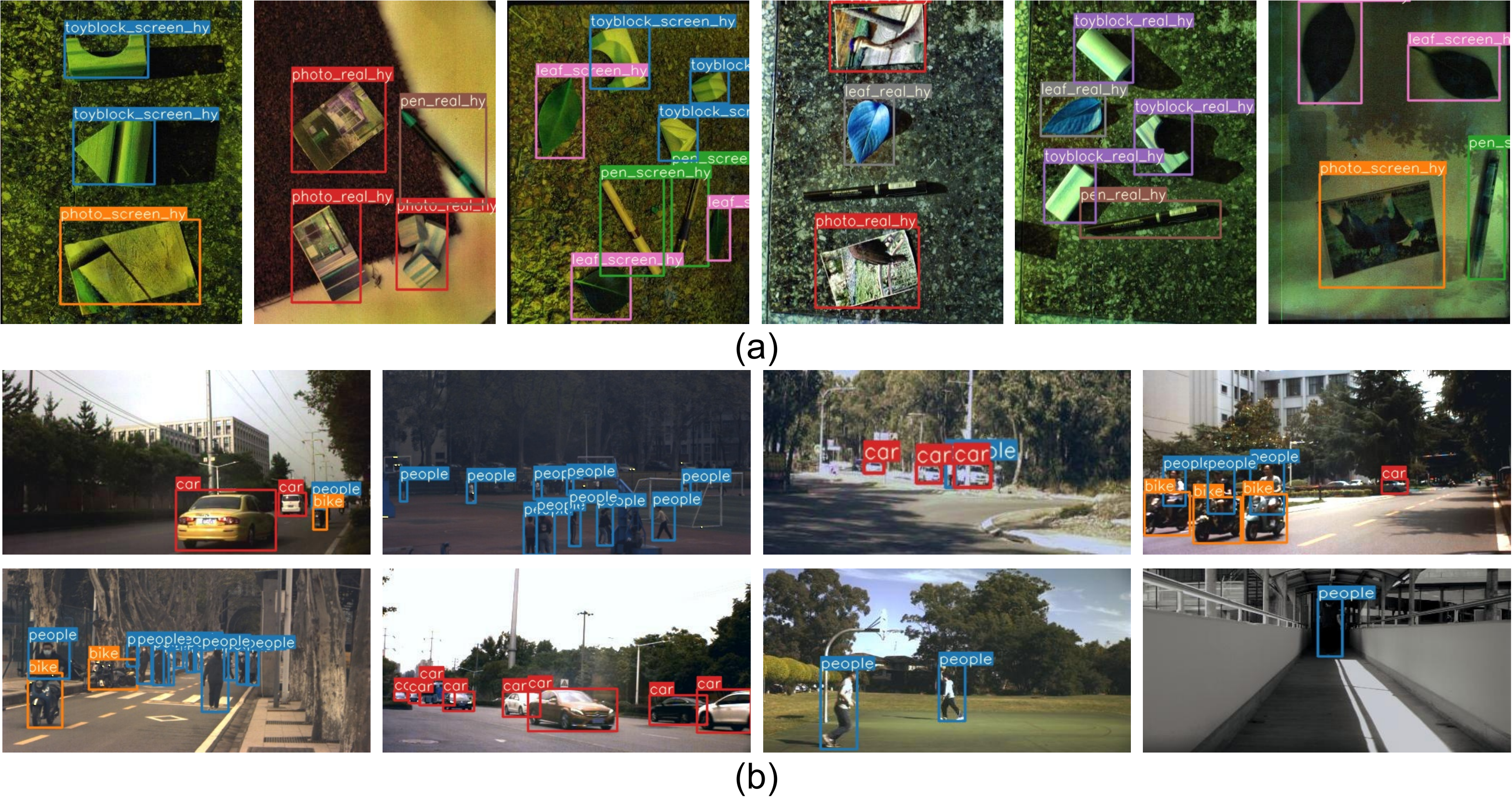}
	\caption {Visualized detection results on SDCM. (a) HOD-1 dataset (b) HOD3K dataset. We convert the hyperspectral images into pseudo-color images for display. The bounding box of a certain color in the image represents the object of the corresponding category.} 	
	\label{img:showtime}
\end{figure*}

\subsection{Discussion}

This section engages in a comprehensive discussion of the diverse phenomena witnessed throughout the experiments, supplemented with insightful suggestions for future research directions.

\subsubsection{Classification Analysis}

In order to provide further evidence of efficacy of the proposed SDCM in enhancing high-level feature discrimination through the mitigation of heterogeneity among hyperspectral bands and the integration of spectral information, we visualized the classification accuracy outcomes of the baseline, S2ADet, and SDCM models, employing the confusion matrix on the HOK-1 dataset.

As illustrated in Fig.~\ref{img:class}, the baseline model struggles to amalgamate information across hyperspectral bands, leading to inconsistent information and inaccurate classification, especially between background and object categories. In contrast, SDCM displays enhanced classification accuracy across all categories. Noteworthy improvements are seen in the pen screen and pen real categories, with classification accuracy more than doubling. Moreover, significant enhancements are observed in the toyblock screen, photo real, toyblock real, and leaf screen categories.

On the other hand, although S2ADet shows some enhancement in classification accuracy, it encounters difficulties in distinguishing between the pen screen and pen real categories. Inspection of detection results (Fig.~\ref{img:showtime}) reveals that the pen category exhibits elongation, providing less distinctive object features and thereby increasing its confusion with the background. In sharp contrast, SDCM effectively addresses this challenge by capitalizing on the spectral information embedded within hyperspectral data to unveil the material's unique characteristics. In comparison to the cutting-edge method (S2ADet), SDCM achieves remarkable improvements of 37\%, 17\%, and 14\% in the pen screen, pen real, and leaf screen categories, respectively. This further reveals the potency of spectral attributes in refining object classification. Hence, the SCL and SDA modules synergistically complement each other, enhancing the comprehension of hyperspectral information.

A scrutiny of Table~\ref{tab:HOD-1} reveals that, for the HOD-1 dataset, SDCM does not consistently achieve peak performance across all categories, particularly in higher-level classification categories. This observation suggests that inter-band consistency learning focuses primarily on capturing vital complementary information while potentially overlooking the discriminative traits unique to a select few bands. Consequently, the feature representation of objects may appear excessively smoothed, ultimately leading to a decline in detection outcomes. Despite potential accuracy diminishment within specific categories, our approach yields substantial improvements across the majority of categories, thereby enhancing the overarching performance.

\subsubsection{Qualitative Analysis}

To further explore the performance of SDCM detector, we visually inspected the detection outcomes of the Baseline, S2ADet, and SDCM detector, as depicted in Fig.~\ref{img:contrast}. The Baseline frequently exhibits missed and erroneous detections in scenes marked by occlusion and overlaps. Conversely, S2ADet tends to confuse objects with the background. In striking comparison, SDCM boasts distinct advantages, inferring object presence with just a few pixels—an achievement that proves intricate with visible images (as evident in the third column of Fig.~\ref{img:contrast}). Therefore, SDCM adeptly harnesses the spectral dimension inherent in hyperspectral images to discern between objects and background, effectively boosting detection accuracy for camouflaged and diminutive objects.

\begin{figure}[t]
    \centering
        \includegraphics[width=0.23\textwidth]{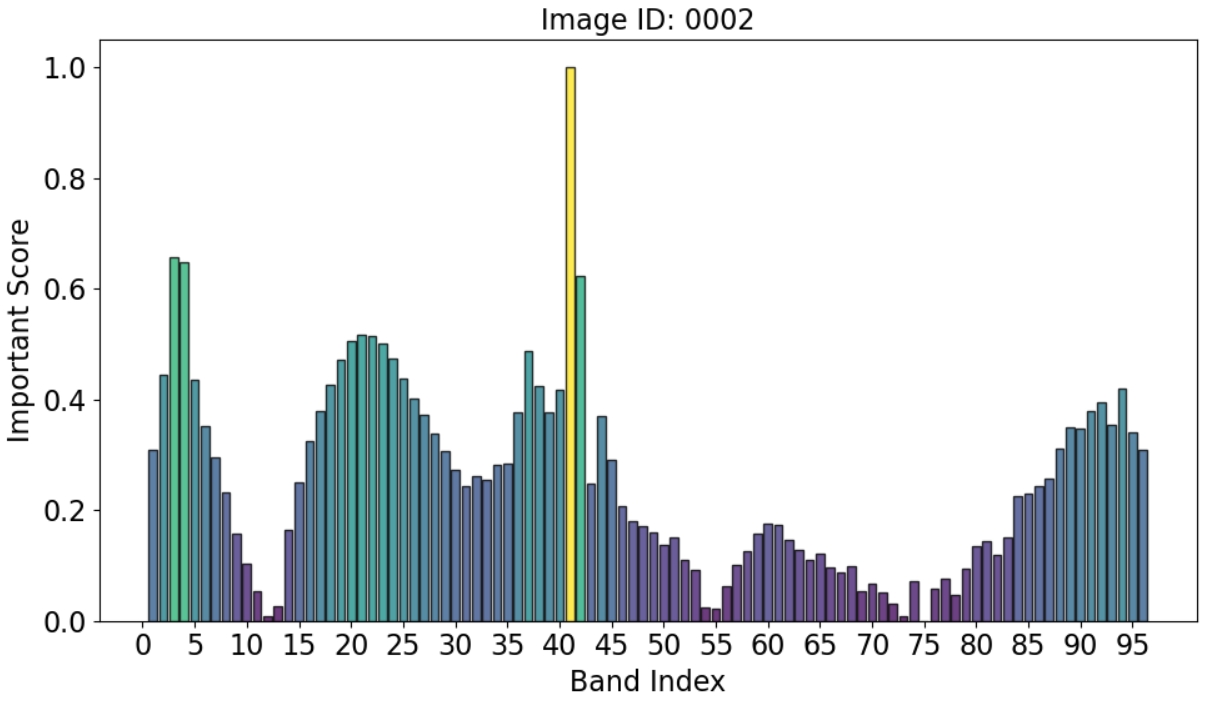}
    \includegraphics[width=0.23\textwidth]{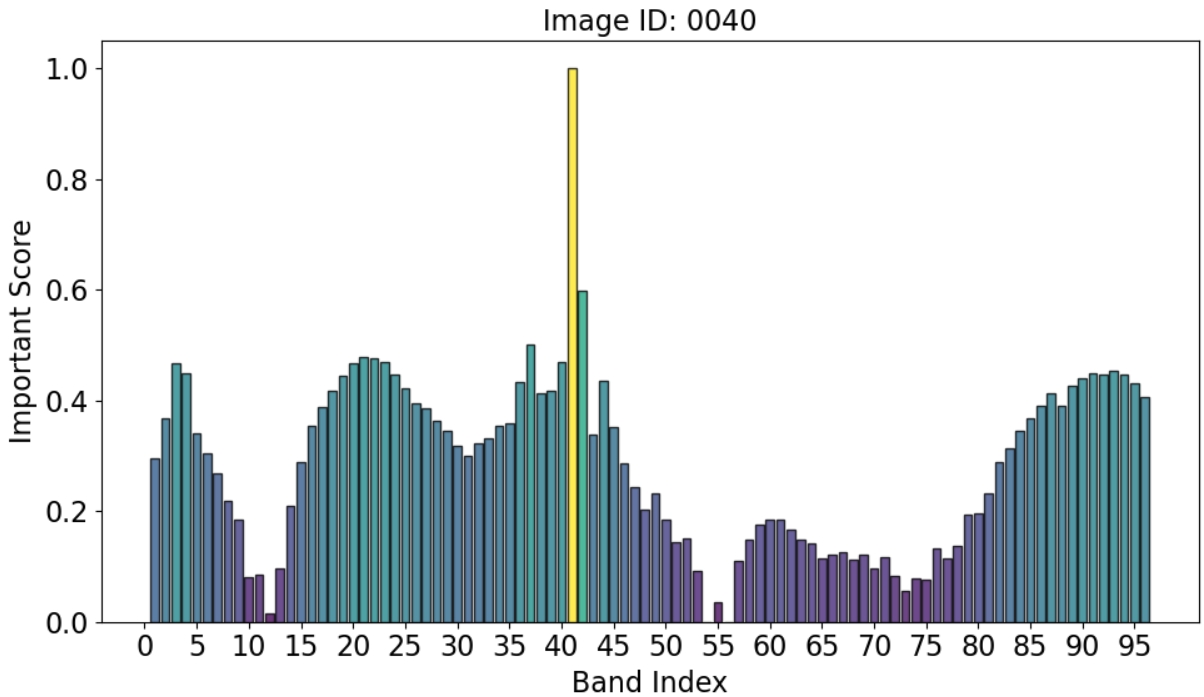}
    \includegraphics[width=0.23\textwidth]{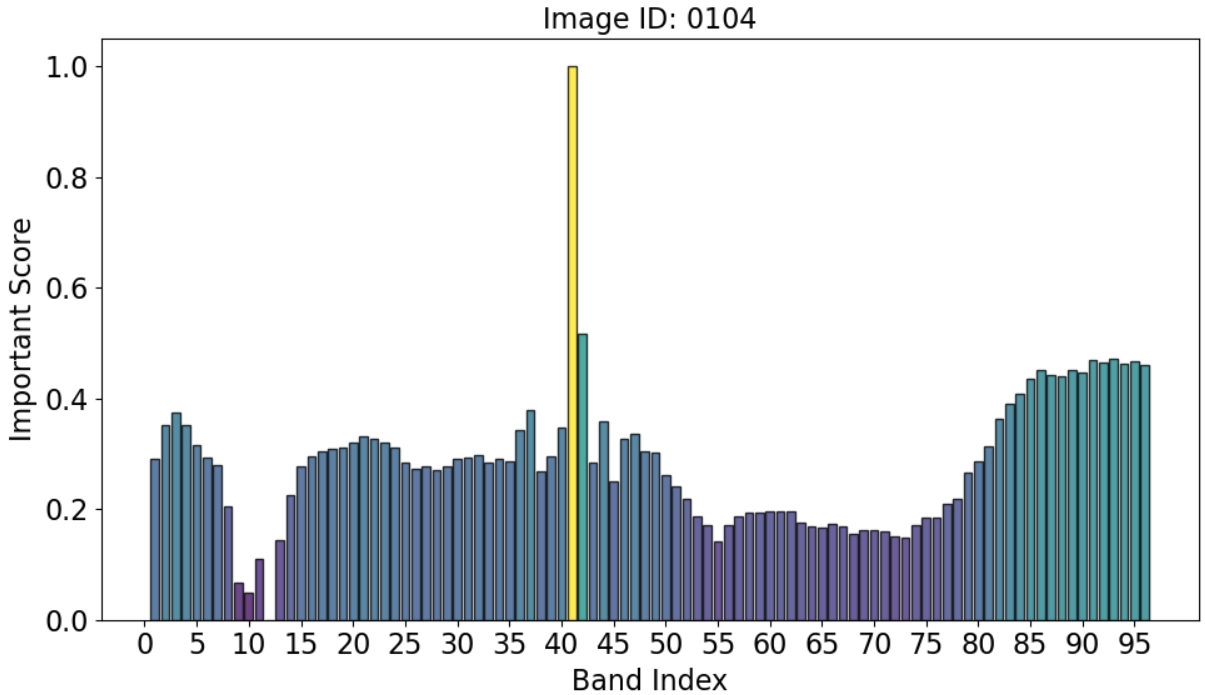}
    \includegraphics[width=0.23\textwidth]{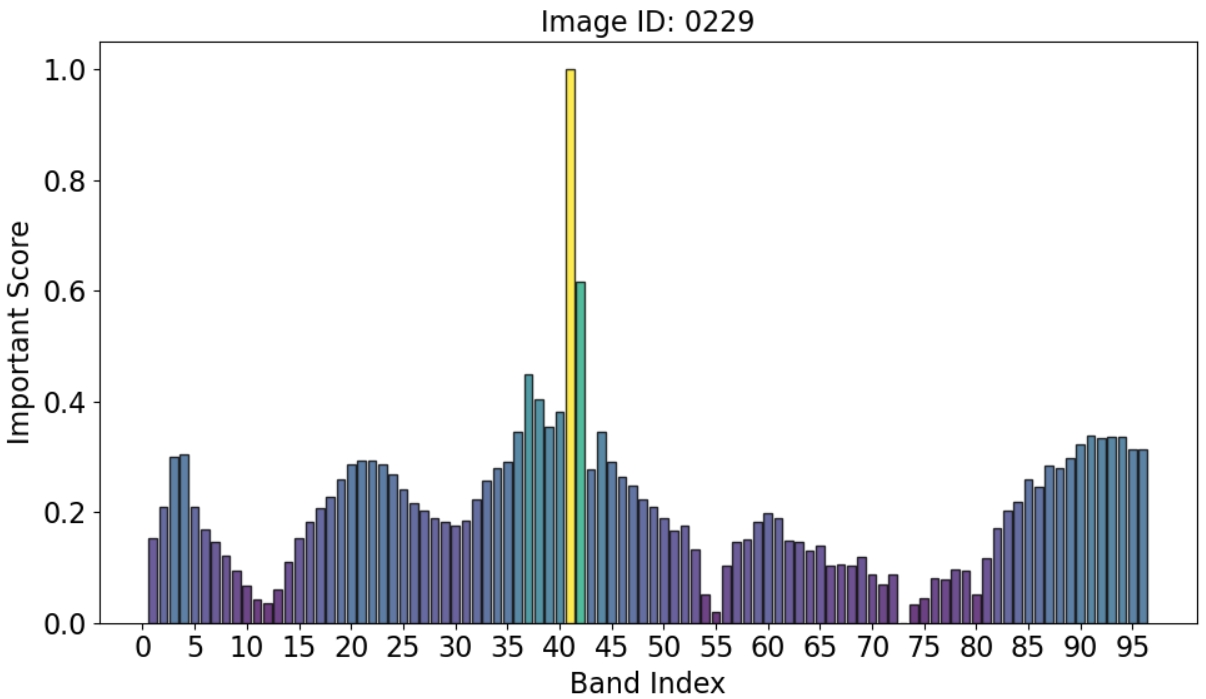}
    \caption{Visualization of three hyperspectral images band importance on the HOD-1 dataset in the SGG Module. A total of 96 bands are in the hyperspectral image.}
    \label{fig:three_images}
\end{figure}

To validate the SGG module, we incorporated a visualization of band importance into the HOD-1 dataset, as shown in Fig.~\ref{fig:three_images}. The importance maps for the three hyperspectral images were normalized to highlight variations in spectral significance. The SGG module effectively isolates key features and selects the most critical spectral bands, enabling a clear and precise assessment of spectral importance.

\subsection{Limitations}

\begin{figure}
	\centering
	\includegraphics[width = 1\linewidth]{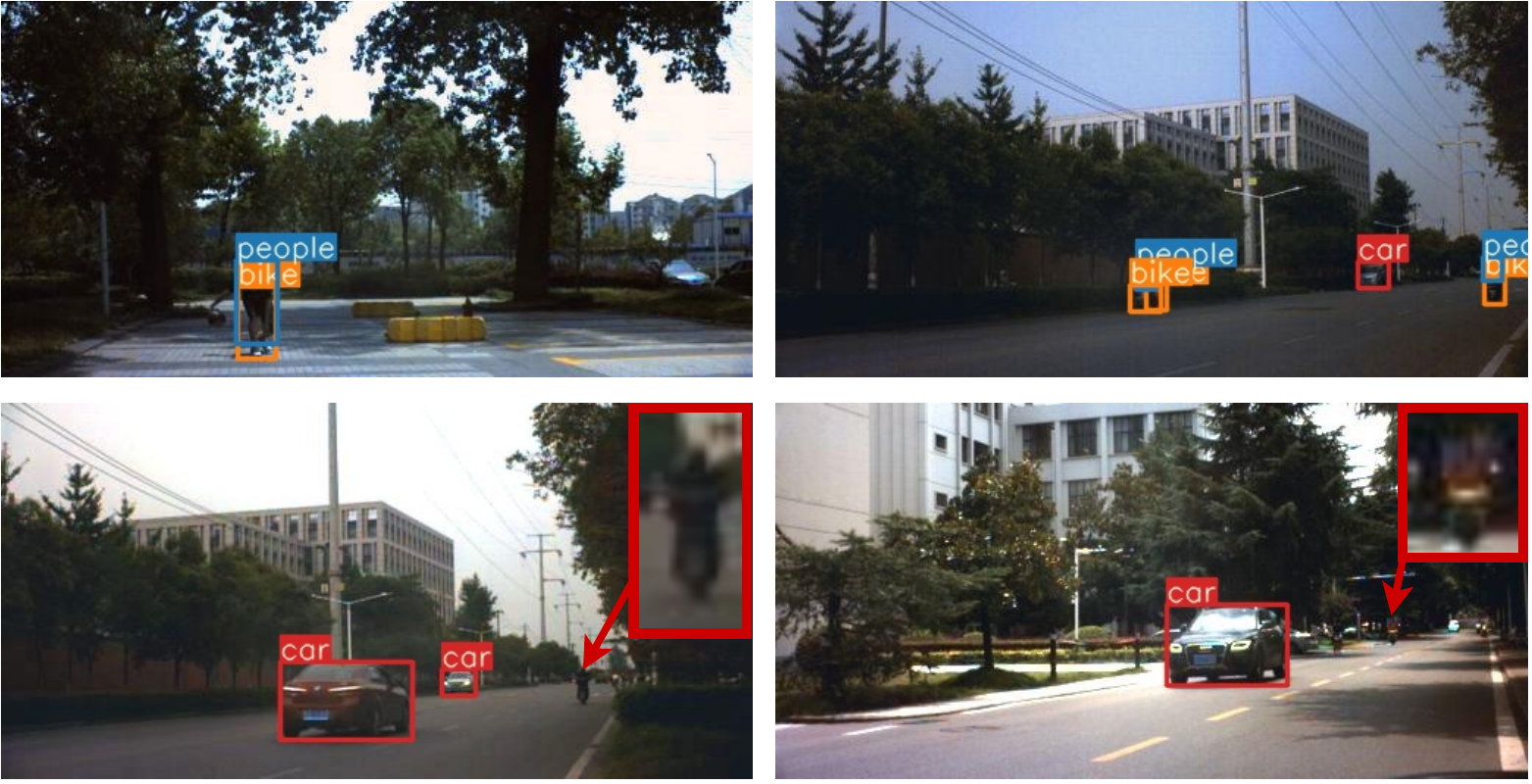}
	\caption {Visualization of some disadvantages of SDCM. The enlarged red box indicates areas of interest.} 	
	\label{img:badshow}
\end{figure}

We discovered certain instances of detection failures through visual examples, as displayed in Fig.~\ref{img:badshow}. Notably, in scenarios involving diminutive objects and occlusion, missed detections frequently arise due to the intrusion of superfluous background information and the original spectral depiction. These objects manifest a relative indistinctness, thereby posing significant challenges for visible image and hyperspectral image analysis. Prospects for optimization lie in leveraging the distinctive attributes of hyperspectral image analysis to ameliorate such limitations. Looking forward, our research endeavors are geared towards the exploration of more potent object detection in hyperspectral images within architectural frameworks, through the seamless integration of spatial and spectral information from diverse wavelength bands into a unified conduit.

\subsection{Visualization Analysis}

The visualization outcomes of SDCM, as illustrated in Fig.~\ref{img:showtime}, also demonstrate its adeptness in accurately identifying camouflaged objects within the HOD-1 dataset. This achievement is attributed to the proficiency of SDCM in fine-tuning object classification for items with akin appearances or shapes, facilitated by the extraction of complementary hyperspectral information via semantic consistency learning. In parallel, SDCM effectively capitalizes on the enriched spectral dimension intrinsic to hyperspectral images, thereby unveiling distinctive material attributes that enable precise differentiation of objects distinguished by intricate details and features.

Fig.~\ref{img:showtime} (b) shows the remarkable prowess of SDCM in scenes replete with dense objects. On the HOD3K dataset, SDCM excels in accurately detecting and localizing multiple closely situated objects, notably individuals. This competence proves invaluable for real-world scenarios demanding the handling of dense object configurations, as encountered in crowd estimation or traffic monitoring applications. Notably, SDCM extends its capabilities to identifying and locating diminutive objects positioned at a distance from the hyperspectral sensor. Despite these objects bearing low resolution and constrained information within the image, SDCM judiciously harnesses the amalgamation of spatial and spectral attributes inherent in hyperspectral analysis. Thus, the visualization outcomes of SDCM, spanning fine-grained classification, dense objects, and diminutive entities, distinctly underscore the algorithm's superior performance within object detection tasks.

\section{Conclusion}
\label{sec:5}
This paper introduces the SDCM, a novel framework designed for the purpose of hyperspectral image object detection. SDCM is adept at managing the inherent heterogeneity existing among hyperspectral bands by leveraging the SCL module, which effectively captures and characterizes the semantic information embedded within hyperspectral data. Furthermore, the SDA architecture seamlessly integrates hyperspectral spectral information, guiding the extraction of high-level features that are crucial for discerning distinct object classes. To curtail redundant hyperspectral information, the SGG module is seamlessly incorporated into the architecture. The central tenet of our approach is the amelioration of inter-band heterogeneity by acquiring consistent spatial information across hyperspectral dimensions through the employment of spatial attention mechanisms. This spatial-spectral fusion, as realized by SDCM, not only facilitates precise object localization but also enhances object classification by revealing material-specific attributes. The results indisputably affirm the superiority and prowess of the SDCM. Remarkably, our approach achieves state-of-the-art detection performance and exhibits substantial enhancements across both the HOD-1 and HOD3K datasets, thus underscoring the robustness of SDCM. Furthermore, the adept SCL and SDA modules empower SDCM to outperform contemporary methods significantly in the realm of object class differentiation.

\section*{Acknowledgment}
The project was supported by the Fundamental Research Funds for National Universities, China University of Geosciences (Wuhan), No.2024XLB7.

\bibliographystyle{IEEEtran}
\bibliography{IEEEabrv,References}

\begin{thebibliography}{10}
\providecommand{\url}[1]{#1}
\csname url@samestyle\endcsname
\providecommand{\newblock}{\relax}
\providecommand{\bibinfo}[2]{#2}
\providecommand{\BIBentrySTDinterwordspacing}{\spaceskip=0pt\relax}
\providecommand{\BIBentryALTinterwordstretchfactor}{4}
\providecommand{\BIBentryALTinterwordspacing}{\spaceskip=\fontdimen2\font plus
\BIBentryALTinterwordstretchfactor\fontdimen3\font minus \fontdimen4\font\relax}
\providecommand{\BIBforeignlanguage}[2]{{%
\expandafter\ifx\csname l@#1\endcsname\relax
\typeout{** WARNING: IEEEtran.bst: No hyphenation pattern has been}%
\typeout{** loaded for the language `#1'. Using the pattern for}%
\typeout{** the default language instead.}%
\else
\language=\csname l@#1\endcsname
\fi
#2}}
\providecommand{\BIBdecl}{\relax}
\BIBdecl

\bibitem{he2023multispectral}
X.~He, C.~Tang, X.~Zou, and W.~Zhang, ``Multispectral object detection via cross-modal conflict-aware learning,'' in \emph{Proceedings of the 31st ACM International Conference on Multimedia}, 2023, pp. 1465--1474.

\bibitem{sun2019hyperspectral}
W.~Sun and Q.~Du, ``Hyperspectral band selection: A review,'' \emph{IEEE Geoscience and Remote Sensing Magazine}, vol.~7, no.~2, pp. 118--139, 2019.

\bibitem{ghamisi2017advances}
P.~Ghamisi, N.~Yokoya, J.~Li, W.~Liao, S.~Liu, J.~Plaza, B.~Rasti, and A.~Plaza, ``Advances in hyperspectral image and signal processing: A comprehensive overview of the state of the art,'' \emph{IEEE Geoscience and Remote Sensing Magazine}, vol.~5, no.~4, pp. 37--78, 2017.

\bibitem{9709645}
L.~Chen, J.~Liu, W.~Chen, and B.~Du, ``A glrt-based multi-pixel target detector in hyperspectral imagery,'' \emph{IEEE Transactions on Multimedia}, vol.~25, pp. 2710--2722, 2023.

\bibitem{liang2018material}
J.~Liang, J.~Zhou, L.~Tong, X.~Bai, and B.~Wang, ``Material based salient object detection from hyperspectral images,'' \emph{Pattern Recognition}, vol.~76, pp. 476--490, 2018.

\bibitem{lu2014medical}
G.~Lu and B.~Fei, ``Medical hyperspectral imaging: a review,'' \emph{Journal of biomedical optics}, vol.~19, no.~1, pp. 010\,901--010\,901, 2014.

\bibitem{bedini2017use}
E.~Bedini, ``The use of hyperspectral remote sensing for mineral exploration: A review,'' \emph{Journal of Hyperspectral Remote Sensing}, vol.~7, no.~4, pp. 189--211, 2017.

\bibitem{lu2020recent}
B.~Lu, P.~D. Dao, J.~Liu, Y.~He, and J.~Shang, ``Recent advances of hyperspectral imaging technology and applications in agriculture,'' \emph{Remote Sensing}, vol.~12, no.~16, p. 2659, 2020.

\bibitem{yan2021object}
L.~Yan, M.~Zhao, X.~Wang, Y.~Zhang, and J.~Chen, ``Object detection in hyperspectral images,'' \emph{IEEE Signal Processing Letters}, vol.~28, pp. 508--512, 2021.

\bibitem{10313066}
P.~Liu, T.~Xu, H.~Chen, S.~Zhou, H.~Qin, and J.~Li, ``Spectrum-driven mixed-frequency network for hyperspectral salient object detection,'' \emph{IEEE Transactions on Multimedia}, pp. 1--15, 2023.

\bibitem{yu2020simplified}
C.~Yu, R.~Han, M.~Song, C.~Liu, and C.-I. Chang, ``A simplified 2d-3d cnn architecture for hyperspectral image classification based on spatial--spectral fusion,'' \emph{IEEE Journal of Selected Topics in Applied Earth Observations and Remote Sensing}, vol.~13, pp. 2485--2501, 2020.

\bibitem{kumar2023methanemapper}
S.~Kumar, I.~Arevalo, A.~Iftekhar, and B.~Manjunath, ``Methanemapper: Spectral absorption aware hyperspectral transformer for methane detection,'' in \emph{Proceedings of the IEEE/CVF Conference on Computer Vision and Pattern Recognition}, 2023, pp. 17\,609--17\,618.

\bibitem{dai2021dynamic}
X.~Dai, Y.~Chen, J.~Yang, P.~Zhang, L.~Yuan, and L.~Zhang, ``Dynamic detr: End-to-end object detection with dynamic attention,'' in \emph{Proceedings of the IEEE/CVF International Conference on Computer Vision}, 2021, pp. 2988--2997.

\bibitem{8762162}
C.~Shi and C.-M. Pun, ``Multiscale superpixel-based hyperspectral image classification using recurrent neural networks with stacked autoencoders,'' \emph{IEEE Transactions on Multimedia}, vol.~22, no.~2, pp. 487--501, 2020.

\bibitem{amenabar2017hyperspectral}
I.~Amenabar, S.~Poly, M.~Goikoetxea, W.~Nuansing, P.~Lasch, and R.~Hillenbrand, ``Hyperspectral infrared nanoimaging of organic samples based on fourier transform infrared nanospectroscopy,'' \emph{Nature communications}, vol.~8, no.~1, p. 14402, 2017.

\bibitem{jiang2022feasibility}
H.~Jiang, W.~Yuan, Y.~Ru, Q.~Chen, J.~Wang, and H.~Zhou, ``Feasibility of identifying the authenticity of fresh and cooked mutton kebabs using visible and near-infrared hyperspectral imaging,'' \emph{Spectrochimica Acta Part A: Molecular and Biomolecular Spectroscopy}, vol. 282, p. 121689, 2022.

\bibitem{yue2022spectral}
J.~Yue, L.~Fang, and M.~He, ``Spectral-spatial latent reconstruction for open-set hyperspectral image classification,'' \emph{IEEE Transactions on Image Processing}, vol.~31, pp. 5227--5241, 2022.

\bibitem{liu2022siamhyper}
Z.~Liu, X.~Wang, Y.~Zhong, M.~Shu, and C.~Sun, ``Siamhyper: Learning a hyperspectral object tracker from an rgb-based tracker,'' \emph{IEEE Transactions on Image Processing}, vol.~31, pp. 7116--7129, 2022.

\bibitem{liu2021anchor}
Z.~Liu, X.~Wang, M.~Shu, G.~Li, C.~Sun, Z.~Liu, and Y.~Zhong, ``An anchor-free siamese target tracking network for hyperspectral video,'' in \emph{2021 11th Workshop on Hyperspectral Imaging and Signal Processing: Evolution in Remote Sensing (WHISPERS)}.\hskip 1em plus 0.5em minus 0.4em\relax IEEE, 2021, pp. 1--5.

\bibitem{sun2020scalability}
P.~Sun, H.~Kretzschmar, X.~Dotiwalla, A.~Chouard, V.~Patnaik, P.~Tsui, J.~Guo, Y.~Zhou, Y.~Chai, B.~Caine \emph{et~al.}, ``Scalability in perception for autonomous driving: Waymo open dataset,'' in \emph{Proceedings of the IEEE/CVF conference on computer vision and pattern recognition}, 2020, pp. 2446--2454.

\bibitem{koz2019ground}
A.~Koz, ``Ground-based hyperspectral image surveillance systems for explosive detection: Part i—state of the art and challenges,'' \emph{IEEE Journal of Selected Topics in Applied Earth Observations and Remote Sensing}, vol.~12, no.~12, pp. 4746--4753, 2019.

\bibitem{wang2022remote}
Y.~Wang, S.~M.~A. Bashir, M.~Khan, Q.~Ullah, R.~Wang, Y.~Song, Z.~Guo, and Y.~Niu, ``Remote sensing image super-resolution and object detection: Benchmark and state of the art,'' \emph{Expert Systems with Applications}, p. 116793, 2022.

\bibitem{ren2015faster}
S.~Ren, K.~He, R.~Girshick, and J.~Sun, ``Faster r-cnn: Towards real-time object detection with region proposal networks,'' \emph{Advances in neural information processing systems}, vol.~28, 2015.

\bibitem{zhao2019object}
Z.-Q. Zhao, P.~Zheng, S.-t. Xu, and X.~Wu, ``Object detection with deep learning: A review,'' \emph{IEEE transactions on neural networks and learning systems}, vol.~30, no.~11, pp. 3212--3232, 2019.

\bibitem{padilla2020survey}
R.~Padilla, S.~L. Netto, and E.~A. Da~Silva, ``A survey on performance metrics for object-detection algorithms,'' in \emph{2020 international conference on systems, signals and image processing (IWSSIP)}.\hskip 1em plus 0.5em minus 0.4em\relax IEEE, 2020, pp. 237--242.

\bibitem{redmon2017yolo9000}
J.~Redmon and A.~Farhadi, ``Yolo9000: better, faster, stronger,'' in \emph{Proceedings of the IEEE conference on computer vision and pattern recognition}, 2017, pp. 7263--7271.

\bibitem{duan2019centernet}
K.~Duan, S.~Bai, L.~Xie, H.~Qi, Q.~Huang, and Q.~Tian, ``Centernet: Keypoint triplets for object detection,'' in \emph{Proceedings of the IEEE/CVF international conference on computer vision}, 2019, pp. 6569--6578.

\bibitem{dai2021up}
Z.~Dai, B.~Cai, Y.~Lin, and J.~Chen, ``Up-detr: Unsupervised pre-training for object detection with transformers,'' in \emph{Proceedings of the IEEE/CVF conference on computer vision and pattern recognition}, 2021, pp. 1601--1610.

\bibitem{guo2022cmt}
J.~Guo, K.~Han, H.~Wu, Y.~Tang, X.~Chen, Y.~Wang, and C.~Xu, ``Cmt: Convolutional neural networks meet vision transformers,'' in \emph{Proceedings of the IEEE Conference on Computer Vision and Pattern Recognition}, 2022, pp. 12\,175--12\,185.

\bibitem{roy2020attention}
S.~K. Roy, S.~Manna, T.~Song, and L.~Bruzzone, ``Attention-based adaptive spectral--spatial kernel resnet for hyperspectral image classification,'' \emph{IEEE Transactions on Geoscience and Remote Sensing}, vol.~59, no.~9, pp. 7831--7843, 2020.

\bibitem{carion2020end}
N.~Carion, F.~Massa, G.~Synnaeve, N.~Usunier, A.~Kirillov, and S.~Zagoruyko, ``End-to-end object detection with transformers,'' in \emph{Computer Vision--ECCV 2020: 16th European Conference, Glasgow, UK, August 23--28, 2020, Proceedings, Part I 16}.\hskip 1em plus 0.5em minus 0.4em\relax Springer, 2020, pp. 213--229.

\bibitem{zhao2022scene}
Y.~Zhao, Y.~Zhang, Z.~Gong, and H.~Zhu, ``Scene representation in bird's-eye view from surrounding cameras with transformers,'' in \emph{Proceedings of the IEEE Conference on Computer Vision and Pattern Recognition}, 2022, pp. 4511--4519.

\bibitem{zhu2020deformable}
X.~Zhu, W.~Su, L.~Lu, B.~Li, X.~Wang, and J.~Dai, ``Deformable detr: Deformable transformers for end-to-end object detection,'' \emph{arXiv preprint arXiv:2010.04159}, 2020.

\bibitem{ge2022mmsrc}
H.~Ge, Y.~Tang, Z.~Bi, T.~Zhan, Y.~Xu, and A.~Song, ``Mmsrc: A multidirection multiscale spectral--spatial residual network for hyperspectral multiclass change detection,'' \emph{IEEE Journal of Selected Topics in Applied Earth Observations and Remote Sensing}, vol.~15, pp. 9254--9265, 2022.

\bibitem{9904945}
D.~Zhu, B.~Du, Y.~Dong, and L.~Zhang, ``Target detection with spatial-spectral adaptive sample generation and deep metric learning for hyperspectral imagery,'' \emph{IEEE Transactions on Multimedia}, vol.~25, pp. 6538--6550, 2023.

\bibitem{7565539}
B.~Du, M.~Zhang, L.~Zhang, R.~Hu, and D.~Tao, ``Pltd: Patch-based low-rank tensor decomposition for hyperspectral images,'' \emph{IEEE Transactions on Multimedia}, vol.~19, no.~1, pp. 67--79, 2017.

\bibitem{fang2023hyperspectral}
L.~Fang, Y.~Jiang, Y.~Yan, J.~Yue, and Y.~Deng, ``Hyperspectral image instance segmentation using spectral--spatial feature pyramid network,'' \emph{IEEE Transactions on Geoscience and Remote Sensing}, vol.~61, pp. 1--13, 2023.

\bibitem{sun2022novel}
H.~Sun, L.~Zhang, J.~Ren, and H.~Huang, ``Novel hyperbolic clustering-based band hierarchy (hcbh) for effective unsupervised band selection of hyperspectral images,'' \emph{Pattern Recognition}, vol. 130, p. 108788, 2022.

\bibitem{ahmad2021hyperspectral}
M.~Ahmad, S.~Shabbir, S.~K. Roy, D.~Hong, X.~Wu, J.~Yao, A.~M. Khan, M.~Mazzara, S.~Distefano, and J.~Chanussot, ``Hyperspectral image classification—traditional to deep models: A survey for future prospects,'' \emph{IEEE journal of selected topics in applied earth observations and remote sensing}, vol.~15, pp. 968--999, 2021.

\bibitem{zhang2023matnet}
B.~Zhang, Y.~Chen, Y.~Rong, S.~Xiong, and X.~Lu, ``Matnet: A combining multi-attention and transformer network for hyperspectral image classification,'' \emph{IEEE Transactions on Geoscience and Remote Sensing}, vol.~61, pp. 1--15, 2023.

\bibitem{liu2019review}
S.~Liu, D.~Marinelli, L.~Bruzzone, and F.~Bovolo, ``A review of change detection in multitemporal hyperspectral images: Current techniques, applications, and challenges,'' \emph{IEEE Geoscience and Remote Sensing Magazine}, vol.~7, no.~2, pp. 140--158, 2019.

\bibitem{chen2014deep}
Y.~Chen, Z.~Lin, X.~Zhao, G.~Wang, and Y.~Gu, ``Deep learning-based classification of hyperspectral data,'' \emph{IEEE Journal of Selected topics in applied earth observations and remote sensing}, vol.~7, no.~6, pp. 2094--2107, 2014.

\bibitem{he2023object}
X.~He, C.~Tang, X.~Liu, W.~Zhang, K.~Sun, and J.~Xu, ``Object detection in hyperspectral image via unified spectral–spatial feature aggregation,'' \emph{IEEE Transactions on Geoscience and Remote Sensing}, vol.~61, pp. 1--13, 2023.

\bibitem{wang2020hyperspectral}
Q.~Wang, F.~Zhang, and X.~Li, ``Hyperspectral band selection via optimal neighborhood reconstruction,'' \emph{IEEE Transactions on Geoscience and Remote Sensing}, vol.~58, no.~12, pp. 8465--8476, 2020.

\bibitem{shukla2016overview}
A.~Shukla and R.~Kot, ``An overview of hyperspectral remote sensing and its applications in various disciplines,'' \emph{IRA-International Journal of Applied Sciences}, vol.~5, no.~2, pp. 85--90, 2016.

\bibitem{yang2021simam}
L.~Yang, R.-Y. Zhang, L.~Li, and X.~Xie, ``Simam: A simple, parameter-free attention module for convolutional neural networks,'' in \emph{International conference on machine learning}.\hskip 1em plus 0.5em minus 0.4em\relax PMLR, 2021, pp. 11\,863--11\,874.

\bibitem{lin2017focal}
T.-Y. Lin, P.~Goyal, R.~Girshick, K.~He, and P.~Doll{\'a}r, ``Focal loss for dense object detection,'' in \emph{Proceedings of the IEEE International Conference on Computer Vision}, 2017, pp. 2980--2988.

\bibitem{tian2019fcos}
Z.~Tian, C.~Shen, H.~Chen, and T.~He, ``Fcos: Fully convolutional one-stage object detection,'' in \emph{Proceedings of the IEEE International Conference on Computer Vision}, 2019, pp. 9627--9636.

\bibitem{wu2020rethinking}
Y.~Wu, Y.~Chen, L.~Yuan, Z.~Liu, L.~Wang, H.~Li, and Y.~Fu, ``Rethinking classification and localization for object detection,'' in \emph{Proceedings of the IEEE Conference on Computer Vision and Pattern Recognition}, 2020, pp. 10\,186--10\,195.

\bibitem{cai2019cascade}
Z.~Cai and N.~Vasconcelos, ``Cascade r-cnn: high quality object detection and instance segmentation,'' \emph{IEEE transactions on pattern analysis and machine intelligence}, vol.~43, no.~5, pp. 1483--1498, 2019.

\bibitem{chen2021you}
Q.~Chen, Y.~Wang, T.~Yang, X.~Zhang, J.~Cheng, and J.~Sun, ``You only look one-level feature,'' in \emph{Proceedings of the IEEE/CVF conference on computer vision and pattern recognition}, 2021, pp. 13\,039--13\,048.

\bibitem{pang2019libra}
J.~Pang, K.~Chen, J.~Shi, H.~Feng, W.~Ouyang, and D.~Lin, ``Libra r-cnn: Towards balanced learning for object detection,'' in \emph{Proceedings of the IEEE Conference on Computer Vision and Pattern Recognition}, 2019, pp. 821--830.

\bibitem{kong2020foveabox}
T.~Kong, F.~Sun, H.~Liu, Y.~Jiang, L.~Li, and J.~Shi, ``Foveabox: Beyound anchor-based object detection,'' \emph{IEEE Transactions on Image Processing}, vol.~29, pp. 7389--7398, 2020.

\bibitem{wold1987principal}
S.~Wold, K.~Esbensen, and P.~Geladi, ``Principal component analysis,'' \emph{Chemometrics and intelligent laboratory systems}, vol.~2, no. 1-3, pp. 37--52, 1987.

\bibitem{he2016deep}
K.~He, X.~Zhang, S.~Ren, and J.~Sun, ``Deep residual learning for image recognition,'' in \emph{Proceedings of the IEEE conference on computer vision and pattern recognition}, 2016, pp. 770--778.

\bibitem{sun2022spectral}
L.~Sun, G.~Zhao, Y.~Zheng, and Z.~Wu, ``Spectral--spatial feature tokenization transformer for hyperspectral image classification,'' \emph{IEEE Transactions on Geoscience and Remote Sensing}, vol.~60, pp. 1--14, 2022.

\bibitem{salvador2021revamping}
A.~Salvador, E.~Gundogdu, L.~Bazzani, and M.~Donoser, ``Revamping cross-modal recipe retrieval with hierarchical transformers and self-supervised learning,'' in \emph{Proceedings of the IEEE/CVF Conference on Computer Vision and Pattern Recognition}, 2021, pp. 15\,475--15\,484.

\bibitem{rezatofighi2019generalized}
H.~Rezatofighi, N.~Tsoi, J.~Gwak, A.~Sadeghian, I.~Reid, and S.~Savarese, ``Generalized intersection over union: A metric and a loss for bounding box regression,'' in \emph{Proceedings of the IEEE/CVF conference on computer vision and pattern recognition}, 2019, pp. 658--666.

\bibitem{redmon2016you}
J.~Redmon, ``You only look once: Unified, real-time object detection,'' in \emph{Proceedings of the IEEE conference on computer vision and pattern recognition}, 2016.

\bibitem{feng2021tood}
C.~Feng, Y.~Zhong, Y.~Gao, M.~R. Scott, and W.~Huang, ``Tood: Task-aligned one-stage object detection,'' in \emph{2021 IEEE/CVF International Conference on Computer Vision (ICCV)}.\hskip 1em plus 0.5em minus 0.4em\relax IEEE Computer Society, 2021, pp. 3490--3499.

\bibitem{redmon2018yolov3}
J.~Redmon and A.~Farhadi, ``Yolov3: An incremental improvement,'' \emph{arXiv preprint arXiv:1804.02767}, 2018.

\bibitem{wang2022kvt}
P.~Wang, X.~Wang, F.~Wang, M.~Lin, S.~Chang, H.~Li, and R.~Jin, ``Kvt: k-nn attention for boosting vision transformers,'' in \emph{European conference on computer vision}.\hskip 1em plus 0.5em minus 0.4em\relax Springer, 2022, pp. 285--302.

\bibitem{lin2014microsoft}
T.-Y. Lin, M.~Maire, S.~Belongie, J.~Hays, P.~Perona, D.~Ramanan, P.~Doll{\'a}r, and C.~L. Zitnick, ``Microsoft coco: Common objects in context,'' in \emph{Computer Vision--ECCV 2014: 13th European Conference, Zurich, Switzerland, September 6-12, 2014, Proceedings, Part V 13}.\hskip 1em plus 0.5em minus 0.4em\relax Springer, 2014, pp. 740--755.

\end{thebibliography}

\end{document}